\documentclass[preprint,10pt,3p]{elsarticle}
\usepackage{geometry}
\usepackage{graphics} 
\usepackage{epsfig} 
\usepackage{times} 
\usepackage{amsmath} 
\usepackage{amssymb}  
\usepackage{epstopdf}
\usepackage{subfigure,color,balance}
\usepackage{caption}
\usepackage{verbatim}
\usepackage{cases}
\allowdisplaybreaks[4]
\usepackage{enumerate}
\usepackage{booktabs}
\usepackage{balance}
\usepackage{mathbbol}
\usepackage{multirow}

\usepackage{algorithm}
\usepackage{algorithmicx}
\usepackage{algpseudocode}
\usepackage{mathrsfs}
\usepackage{threeparttable}
\usepackage[draft]{todonotes}
\usepackage{makecell}

\usepackage{enumitem}

\usepackage[colorlinks=false,      
linkcolor=black,      
citecolor=black,      
filecolor=black,      
urlcolor=blue]{hyperref}

\newtheorem{theorem}{Theorem}

\newdefinition{remark}{Remark}
\newtheorem{lemma}{Lemma}

\newtheorem{problem}{Problem}

\def\x{{\bf x}}

\def\0{{\bf 0}}
\def\1{{\bf 1}}

\usepackage{wrapfig}

 \biboptions{authoryear}

\begin{document}
	\begin{frontmatter}
		\title{Trustworthy Human-AI Collaboration: Reinforcement Learning with Human Feedback and Physics Knowledge for Safe Autonomous Driving}

            \author{Zilin Huang}
		\ead{zilin.huang@wisc.edu}
		\author{Zihao Sheng}
		\ead{zihao.sheng@wisc.edu}
		\author{Sikai Chen\corref{cor1}}
		\ead{sikai.chen@wisc.edu}

		\cortext[cor1]{Corresponding author: Sikai Chen}
  
		\address{Department of Civil and Environmental Engineering, University of Wisconsin-Madison, Madison, WI, 53706, USA}
		
\begin{abstract}
In the field of autonomous driving, developing safe and trustworthy autonomous driving policies remains a significant challenge. Recently, Reinforcement Learning with Human Feedback (RLHF) has attracted substantial attention due to its potential to enhance training safety and sampling efficiency. Nevertheless, existing RLHF-enabled methods often falter when faced with imperfect human demonstrations, potentially leading to training oscillations or even worse performance than rule-based approaches. Inspired by the human learning process, we propose \textbf{Physics-enhanced Reinforcement Learning with Human Feedback (PE-RLHF)}. This novel framework synergistically integrates human feedback (e.g., human intervention and demonstration) and physics knowledge (e.g., traffic flow model) into the training loop of reinforcement learning. The key advantage of PE-RLHF is that the learned policy will perform at least as well as the given physics-based policy, even when human feedback quality deteriorates, thus ensuring trustworthy safety improvements. PE-RLHF introduces a Physics-enhanced Human-AI (PE-HAI) collaborative paradigm for dynamic action selection between human and physics-based actions, employs a reward-free approach with a proxy value function to capture human preferences, and incorporates a minimal intervention mechanism to reduce the cognitive load on human mentors. Extensive experiments across diverse driving scenarios demonstrate that PE-RLHF significantly outperforms traditional methods, achieving state-of-the-art (SOTA) performance in safety, efficiency, and generalizability, even with varying quality of human feedback. The philosophy behind PE-RLHF not only advances autonomous driving technology but can also offer valuable insights for other safety-critical domains. Demo
video and code are available at: \href{https://zilin-huang.github.io/PE-RLHF-website/}{\textcolor{magenta}{https://zilin-huang.github.io/PE-RLHF-website/}}.

\end{abstract}
		
		
	\end{frontmatter}
	
\section{Introduction}

Autonomous driving technology holds significant potential to enhance traffic safety and mobility across various driving scenarios \citep{feng2023dense,cao2023continuous,huang2024toward}. Several autonomous vehicles (AVs) companies have recently demonstrated impressive performance metrics. For instance, in 2023, Waymo's AVs traveled a total of 4,858,890 miles in California. Similarly, Cruise's AVs achieved 2,064,728 driverless miles and 583,624 miles with a safety driver, while Zoox reported 710,409 miles with a safety driver and 11,263 miles without one in California \citep{robotreport2023}. Despite these advancements, autonomous driving technology remains far from achieving full automation (Level 5) across all driving scenarios \citep{SAE_J3016_2021}. In particular, developing safe and generalizable driving policies for various safety-critical scenarios remains an ongoing research challenge. \citep{lin2024safety,huang2024human,mao2024integrating}. A recent survey highlighted that safety, rather than economic consequences or privacy issues, is the primary concern regarding AV acceptability \citep{ju2022survey}. Moreover, various agencies and the public still harbor concerns about the trustworthiness of autonomous driving systems \citep{cao2022trustworthy,cao2023continuous,he2024toward,he2024trustworthy}. Therefore, it is imperative to bridge the gap between the anticipated autonomous future and the current state-of-the-art technology by developing trustworthy, safety-assured driving policies.

\begin{figure*}[t]
\centering
  \includegraphics[width=0.9999\textwidth]{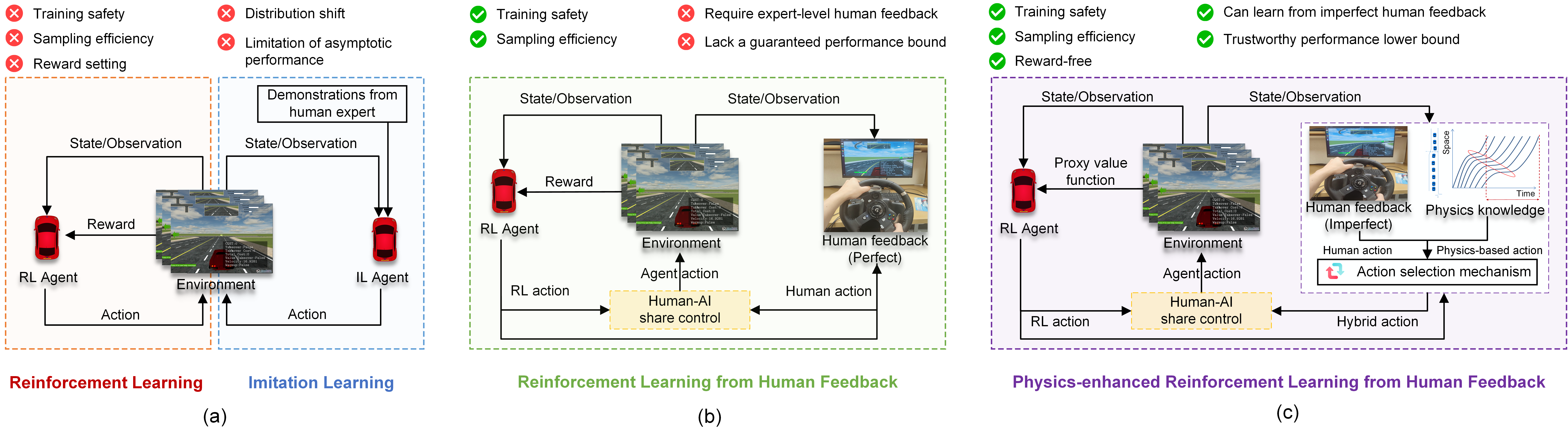}
  \caption{Motivation for this work. (a) Fundamentals and limitations of existing IL/RL-based methods for driving policy learning. (b) Fundamentals and limitations of RLHF-based methods. (c) Fundamentals of our proposed PE-RHLF framework, which achieves trustworthy safety improvements.}
  \label{fig1}
\end{figure*}

Generally, AV companies employ a hierarchical approach to decompose the driving task into multiple sub-tasks. This approach reduces computational complexity and provides good decision-making transparency. Nevertheless, it requires cumbersome hand-crafted rules and may fail in difficult and highly interactive scenarios \citep{cao2021confidence, yang2023towards, wu2024human}. In recent years, learning-based end-to-end approaches have attracted increasing attention since they can learn from collected driving data, offering a potential path for designing more efficient driving policies \citep{you2024v2x,sheng2024ego}. A notable example is UniAD, the 2023 CVPR Best Paper \citep{hu2023planning}. As shown in Fig. \ref{fig1} (a), imitation learning (IL) and reinforcement learning (RL) are two main approaches, particularly in the context of end-to-end driving policy learning \citep{huang2024human}. IL aims to learn driving policies by mimicking human driving behavior. While IL has shown good performance in specific decision-making scenarios, it faces two significant issues in practical applications: distribution shift and limitations in asymptotic performance \citep{wu2023toward}. \cite{ding2023survey} illustrates that even minor domain shifts in road structures or surrounding vehicles can lead to catastrophic outcomes due to the high-stakes nature of autonomous driving.  

RL leverages iterative self-improvement, offering the potential to mitigate inherent limitations associated with imitation-based approaches. The effectiveness of RL-enabled methods has been demonstrated in various decision-making scenarios, such as  highway exiting \citep{chen2021graph}, traffic congestion \citep{sheng2024traffic}, and lane-changing \citep{li2022decision}. Nevertheless, RL typically requires extensive interactions with the environment, which can reduce sampling efficiency and raise safety concerns during both the training and testing phases \citep{wu2024recent}. Additionally, designing an appropriate reward function to capture all desired driving behavior can be challenging \citep{peng2024learning}. If not carefully crafted, these may lead to unintended consequences. \cite{knox2023reward} found that many of the reward functions proposed in the autonomous driving literature failed basic consistency checks, which could lead to unsafe behavior. Consequently, few AV companies are ready to deploy this technique in their production AVs \citep{cao2023continuous}. Some studies have sought to improve the trustworthiness of RL by improving expected performance, e.g., cost constraints \citep{stooke2020responsive} or action safeguards \citep{yang2023towards}. Yet, these methods still face challenges in ensuring safety, interpretability, and sampling efficiency. 

Recently, researchers in the computer science community have explored incorporating human knowledge into the RL learning processes, leading to the development of Reinforcement Learning from Human Feedback (RLHF) \citep{christiano2017deep}.  This method leverages humans' superior adaptability, situational awareness, and decision-making skills in contextual understanding and knowledge-based reasoning \citep{huang2024human}.  Different from traditional RL, RLHF bypasses the complex task of reward design by directly training a `reward model' from human feedback. Technically, RLHF aims to align agents with human preferences rather than merely mimicking human behavior. Despite being a relatively young technique, RLHF has garnered significant public attention through high-profile AI applications, including large language models (e.g., ChatGPT \citep{openai2022chatgpt}), home robots (e.g., Mobile ALOHA \citep{fu2024mobile}), and text-to-image models (e.g., Midjourney \citep{midjourney}). The RLHF approach has proven effective in enhancing both the training safety and sampling efficiency in RL. Some studies have attempted to apply the philosophy of the RLHF technique to the autonomous driving domain, specifically in decision-making scenarios such as safe navigation \citep{wu2023toward,li2022efficient}, lane-changing \citep{wu2024human,huang2023human,wang2024reinforcement}, trajectory planning \citep{zhuang2024hgrl}, and unprotected left turn \citep{wu2022prioritized,huang2022efficient}. As shown in Fig. \ref{fig1} (b), humans can effortlessly adapt their driving behaviors based on static contexts such as roadblocks, or dynamic contexts such as surrounding traffic, often making intuitive judgments.

While these studies have achieved many competitive results, two significant challenges remain. First, most studies require expert-level human feedback. In reality, however, consistently obtaining high-quality human feedback is time-consuming and, at times, nearly impossible \citep{huang2024safety}. This is because there is no theoretical guarantee that each human participant has the requisite driving proficiency. It is equally possible that even the experts could perform degraded or catastrophic actions due to various factors, such as distractions or fatigue from prolonged interactions \citep{lin2024safety}. Second, until now, few studies have focused on the trustworthiness of RLHF-enabled methods. While some recent works have begun to address the issue of imperfect human feedback, for example, by incorporating environmental information \citep{huang2024safety} or adaptive weighting factors \citep{wu2023toward}, the generated policies still lack a guaranteed performance bound for driving. In the real world, human learning skills often rely on both human teachers and established knowledge. For instance, as shown in Fig. \ref{fig2} (a), when learning a foreign language, we consult native speakers for practical usage and textbooks for grammatical rules. When native speakers' explanations are unclear, learners can refer to grammar books as a reliable reference. Similarly, in transportation science, there are many well-established physics knowledge (e.g., traffic flow model \citep{treiber2000congested,kesting2007general} ). These models have been widely applied in various aspects of transportation engineering, including traffic management and advanced driver assistance systems (ADAS). 

These observations motivate us to explore a crucial question: \textit{Can we develop a novel interactive learning scheme that allows RL to simultaneously learn from both human feedback and physics knowledge, thereby ensuring `trustworthy safety improvement' in RLHF techniques?} In this work, we propose a novel framework named  \textbf{Physics-enhanced Reinforcement Learning with Human Feedback (PE-RLHF)} to bridge the aforementioned gap. As shown in Fig. \ref{fig1} (c), the uniqueness of PE-RLHF lies in its ability to maintain the benefits of RLHF in enhancing training safety and sampling efficiency, while leveraging physics knowledge to mitigate training oscillations or divergence caused by imperfect human feedback. In other words, we relax the assumption of perfect human mentors to account for situations where humans occasionally provide sub-optimal demonstration.  The concept of `trustworthy' in this work implies that the proposed PE-RLHF establishes a performance floor, guaranteed by a controllable and interpretable physics-based model, even when human feedback quality deteriorates. We emphasize that this work does not aim to solve all trustworthiness issues in autonomous driving, but rather provides a framework that can combine the advantages of RLHF and physics-based methods for achieving better safety performance (i.e., trustworthy improvements).  

The main contributions of our work are as follows: \footnote{Code is available at: \href{https://github.com/zilin-huang/PE-RLHF}{\textcolor{magenta}{https://github.com/zilin-huang/PE-RLHF}}}

\begin{itemize}
\item Inspired by human learning processes, we design a novel Physics-enhanced Human-AI (PE-HAI) collaborative paradigm that ensures a trustworthy safety performance lower bound even when human feedback quality deteriorates. Specifically, we develop an action selection mechanism that dynamically chooses between human and physics-based actions, guaranteeing that the agent executes the action with the higher policy value. 
\item Building upon PE-HAI, we propose PE-RLHF, which, to the best of our knowledge, is the first framework that synergistically integrates human feedback (e.g., human intervention and demonstration) with physics knowledge (e.g., traffic flow model) into the RL training loop for driving policy learning.  We provide theoretical guarantees of the performance improvement of PE-RLHF over existing physics-based policies. 
\item We employ a reward-free approach with a proxy value function to represent human preferences and guide the training process, thus circumventing the challenges of reward design. To enhance the accuracy and robustness of value estimation, we implement an ensemble Q-network technique. Additionally, we incorporate a minimal intervention mechanism to reduce the cognitive load on human mentors. 
\item We conduct extensive experiments across various driving scenarios characterized by high uncertainty and complexity. The results demonstrate the superior and robust performance of PE-RLHF in terms of training safety, sampling efficiency, and generalizability, even when faced with varying qualities of human feedback, compared to state-of-the-art (SOTA) methods. 
\end{itemize}

The remainder of this paper is organized as follows. Section \ref{Related Works} provides a comprehensive review of related work. Section \ref{Preliminaries and Problem Formulation} introduces the problem formulation. Section \ref{Physics-enhanced Human-AI Collaborative Paradigm} presents the proposed PE-HAI collaborative paradigm. Section \ref{Physics-enhanced Reinforcement Learning with Human Framework} describes the proposed PE-RLHF framework. Section \ref{Experimental Evaluation} presents extensive experimental evaluations, comparing PE-RLHF with SOTA methods under different conditions. Finally, Section \ref{Conclusions and Future Work} concludes the paper with a summary of findings and discusses future research directions.

\section{Related Works}
\label{Related Works}

\subsection{Safety Guarantees of RL-based Decision-Making}
The trial-and-error nature of RL exposes agents to potentially dangerous situations, thus limiting its applicability in the domain of autonomous driving \citep{wu2023toward,wu2024recent,li2022efficient,peng2022safe}. Numerous studies have attempted to develop safety guarantee techniques to ensure both training and testing safety, which can be broadly categorized into three approaches: safe RL, offline RL, and action safeguards. Safe RL aims to ensure that each updated policy meets specified constraints. These methods often require knowledge of the probability that a policy will violate constraints. Various constraint optimization methods have been explored, such as trust region methods (e.g., constrained policy optimization, CPO \citep{achiam2017constrained}) and Lagrangian methods (e.g., PPO-Lag \citep{stooke2020responsive} and SAC-Lag \citep{ha2021learning}). Nevertheless, insufficient data or inaccurate models can lead to significant errors, potentially causing these methods to fail. Offline RL methods, such as conservative Q-Learning \citep{kumar2020conservative}, aim to learn conservative policies from pre-collected datasets without online interaction, thereby avoiding potential safety risks during exploration \citep{levine2020offline}. However, offline RL methods may struggle with generalization to unseen scenarios since they can not utilize online exploration data.

The fundamental idea of action safeguards is to combine RL-based policies with physics-based policies. When an action generated by the RL policy is deemed dangerous, the physics-based policy is employed instead. The detection of dangerous actions can be designed based on model uncertainty \citep{yang2023towards}, policy confidence \citep{cao2021confidence,cao2023continuous}, or driving risk estimations \citep{cao2022trustworthy,bai2024longitudinal}. While they can enhance the safety of autonomous driving systems, these methods can sometimes be overly conservative for practical driving. Furthermore, they do not address the fundamental issue of low sample efficiency in RL. In this work, we propose a novel PE-RLHF framework that combines human feedback with physics-based safeguards. Our main insight is that human mentors can provide safe and efficient actions in most driving scenarios. This insight allows us to leverage human feedback to guide the RL agent's learning process, ensuring safety while significantly reducing the number of interactions required to learn effective policies.

\subsection{RLHF for Driving Policy Learning}
In RLHF, human feedback can be broadly categorized into three types: evaluation (such as ranking or rating), intervention, and demonstration \citep{huang2024human}. Evaluation-based methods typically involve humans assessing trajectories sampled by the learning agent or advising on actions when requested by the agent \citep{christiano2017deep,wang2024reinforcement}.  This passive human involvement can pose a risk to the human-AI collaborative system, as the agent explores the environment without sufficient safeguards \citep{peng2022safe}. Demonstration-based methods learn from collected offline and static demonstration data. The agent treats the demonstrations as optimal transitions to imitate. If the provided data lacks reward signals, the agent can learn by imitating the teacher's policy distribution, matching trajectory distributions, or reshaping the expert's reward \citep{ly2020learning,wu2022prioritized,zhuang2024hgrl}. With additional reward signals, the agent can perform pessimistic Bellman updates, similar to most offline RL methods \citep{levine2020offline}. Intervention-based methods allow human mentors to intervene in the training process. The switching between the expert's and the RL policy can be rule-based \citep{peng2022safe}, or decided by the expert \citep{li2022efficient, wu2023toward}.  \cite{li2022efficient} proves safety guarantees in intervention-based methods, providing an additional lower bound on cumulative rewards. Nevertheless, many existing studies still adhere to the traditional RL paradigm, which necessitates the design of a reward function \citep{wu2023toward,huang2023human}.  Manually designing a reward function that effectively encompasses all driving behaviors is a significant challenge. Meanwhile, only a few works consider constraints to reduce human intervention \citep{li2022efficient}. 

In our previous work \citep{huang2024human}, we proposed a human as AI mentor-based deep RL framework (HAIM-DRL), which simultaneously considered a reward-free setting and reduced the cognitive load of human mentors.  However, to the best of our knowledge, most RLHF-enabled methods for driving policy learning have not considered trustworthy safety guarantees in the case of imperfect human feedback.  Perhaps the closest to this work is  \cite{wu2023toward}, which proposes an adaptive weighting factor to adjust the credibility of human actions by evaluating the potential advantages of human behavior relative to the RL policy. While this method can relax the requirements for the quality of human demonstrations, the generated policy still lacks a performance bound for driving. In this work, we propose a novel PE-RLHF framework that integrates both human feedback and physics knowledge to address these limitations.  Moreover, \cite{huang2024safety} proposes a safety-aware human-in-the-loop reinforcement learning approach to alleviate the risk of human feedback deterioration. While their motivation is similar, they achieve this by incorporating environmental information,  whereas we take advantage of the trustworthy performance bound provided by physics knowledge while maintaining the adaptability and efficiency of human feedback.

\section{Problem Formulation}
\label{Preliminaries and Problem Formulation}

\subsection{Preliminaries}
The driving policy learning problem can be formulated as a Markov decision process (MDP) \citep{cao2021confidence}. An MDP is defined by a tuple $(S, A, P, R, \mu, \gamma)$, where $S$ and $A$ denote the state space and action space, respectively \citep{sutton2018reinforcement}. In the context of autonomous driving, the state space may include information about the ego vehicle, surrounding vehicles, and the driving environment, while the action space may consist of control inputs such as throttle and steering. The transition probability function $P : S \times A \times S \mapsto [0, 1]$ describes the dynamics of the system, and the reward function $R : S \times A \mapsto \mathbb{R}$ encodes the desired driving behavior. The initial state distribution is denoted by $\mu : S \mapsto [0, 1]$, and $\gamma$ is the discount factor for future rewards.

The goal of standard RL is to find the optimal policy $\pi^*$ that maximizes the expected discounted return $J_R(\pi)$, which is defined as:
\begin{equation}
\label{eq1}
J_R(\pi) = \mathbb{E}_{\tau \sim \pi}\left[\sum_{t=0}^{\infty} \gamma^t R(s_t, a_t)\right]
\end{equation}
where $\tau = \{(s_t, a_t)\}_{t \geq 0}$ represents a sample trajectory, and $\tau \sim \pi$ denotes the distribution over trajectories generated by policy $\pi$. The distribution of the initial state $s_0$ is given by $\mu$, the action $a_t$ is sampled from the policy $\pi(\cdot | s_t)$ at each time step $t$, and the next state $s_{t+1}$ is sampled from the transition probability function $P(\cdot | s_t, a_t)$.

\subsection{Problem Statement}
In this work, our goal is to develop a safe and trustworthy driving policy learning framework for autonomous driving. This framework should be characterized as follows: (a) It should be able to provide trustworthy safety performance guarantees. (b) It should have strong generalization ability for environmental uncertainties of real-world traffic scenarios (e.g., changing road geometries and unforeseen obstacles). (c) It should have high sampling efficiency with limited training data. To realize this goal, we propose integrating human feedback and physics knowledge from traffic science into the training loop of the RL. By incorporating human feedback, the agent can learn with higher sampling efficiency. Furthermore, by incorporating physics knowledge, we can statistically guarantee better safety performance than a given physics-based policy.  In detail, we decompose the goal into the following sub-problems:

\begin{problem}[Learning from Human Feedback]
\label{problem1}
Traditional learning-based methods, such as IL and RL, face challenges in ensuring safety and sampling efficiency. To take advantage of human intelligence, we should design a scheme that enables AV agents to learn from human feedback (e.g., intervention and demonstration).
\end{problem}

We define a dataset of human demonstration $\mathcal{D}_{\text{human}} = \{(s_t, a_t^{\text{human}})\}$, where $s_t$ represents the state and $a_t^{\text{human}} \sim \pi_{\text{human}}(\cdot \mid s_t)$ represents the action taken by the human in state $s_t$. Then, we aim to train an AV agent with policy $\pi_{\text{AV}}$ from $\mathcal{D}_{\text{human}}$ that can make wise decisions $a_t^{\text{AV}}$ given state $s_t$. In other words, we need to align its behavior with human preferences as closely as possible:
\begin{equation}
\label{eq4}
\pi_{\text{AV}}^* = \arg\min_{\pi_{\text{AV}}} \mathbb{E}_{s_t \sim d_{\pi_{\text{AV}}}} \left[\mathcal{L}\left(\pi_{\text{AV}}(\cdot | s_t), \pi_{\text{human}}(\cdot | s_t)\right)\right]
\end{equation}
where $d_{\pi_{\text{AV}}}$ represents the state distribution induced by the agent's policy $\pi_{\text{AV}}$, and $\mathcal{L}(\cdot, \cdot)$ is a measure of discrepancy between the agent's policy and the human policy, such as the KL divergence \citep{huang2022efficient}. By minimizing the discrepancy between the agent's policy and the human policy over the state distribution, we encourage the agent to align its behavior with human preferences.  

\begin{problem}[Trustworthy Safety Improvement]
\label{problem2}
Due to factors such as limited perception, distraction, or fatigue, the quality of human demonstration may decline over time, leading to training failure. To ensure the effectiveness and trustworthiness of RLHF-enabled methods, we should guarantee that even when the quality of human demonstration decreases, the performance of the AV agent's policy is still not inferior to existing physics-based methods.
\end{problem}

Inspired by the action safeguards methods, we can leverage well-established physics-based models (capable of handling most driving cases except for long-tail scenarios) from traffic science as a trustworthy lower bound for the safety performance of the AV agent's policy. Formally, we define the problem as follows: 
\begin{equation}
\label{eq5}
\begin{aligned}
\pi_{\text{AV}}^* & = \arg\min_{\pi_{\text{AV}}} \mathbb{E}_{s_t \sim d_{\pi_{\text{AV}}}} \left[\mathcal{L}\left(\pi_{\text{AV}}(\cdot | s_t), \pi_{\text{hybrid}}(\cdot | s_t)\right)\right] \\
\text{s.t.} \quad \mathbb{E}_{\tau \sim \pi_{\text{hybrid}}}\left[\sum_{t=0}^{H} \gamma^t r(s_t, a_t)\right] & = \max\left(\mathbb{E}_{\tau \sim \pi_{\text{human}}}\left[\sum_{t=0}^{H} \gamma^t r(s_t, a_t)\right], \mathbb{E}_{\tau \sim \pi_{\text{phy}}}\left[\sum_{t=0}^{H} \gamma^t r(s_t, a_t)\right]\right) \\
& \geq \mathbb{E}_{\tau \sim \pi_{\text{phy}}}\left[\sum_{t=0}^{H} \gamma^t r(s_t, a_t)\right]
\end{aligned}
\end{equation}
where $H$ is the planning horizon.  The hybrid policy $\pi_{\text{hybrid}}$ is defined as the policy that selects the action with the higher expected return between the human policy $\pi_{\text{human}}$ and the physics-based policy $\pi_{\text{phy}}$. This formulation ensures that the AV agent's performance is at least as good as the $\pi_{\text{phy}}$, while allowing it to benefit from human feedback when available. Importantly, this approach provides a trustworthy safety lower bound if the quality of human demonstration deteriorates. By learning from $\pi_{\text{phy}}$, the AV agent can potentially surpass the performance of both the $\pi_{\text{human}}$  and $\pi_{\text{phy}}$. 

\section{Physics-enhanced Human-AI Collaborative Paradigm}
\label{Physics-enhanced Human-AI Collaborative Paradigm}

\subsection{Inspiration}
\begin{figure*}[t]
\centering
  \includegraphics[width=0.9\textwidth]{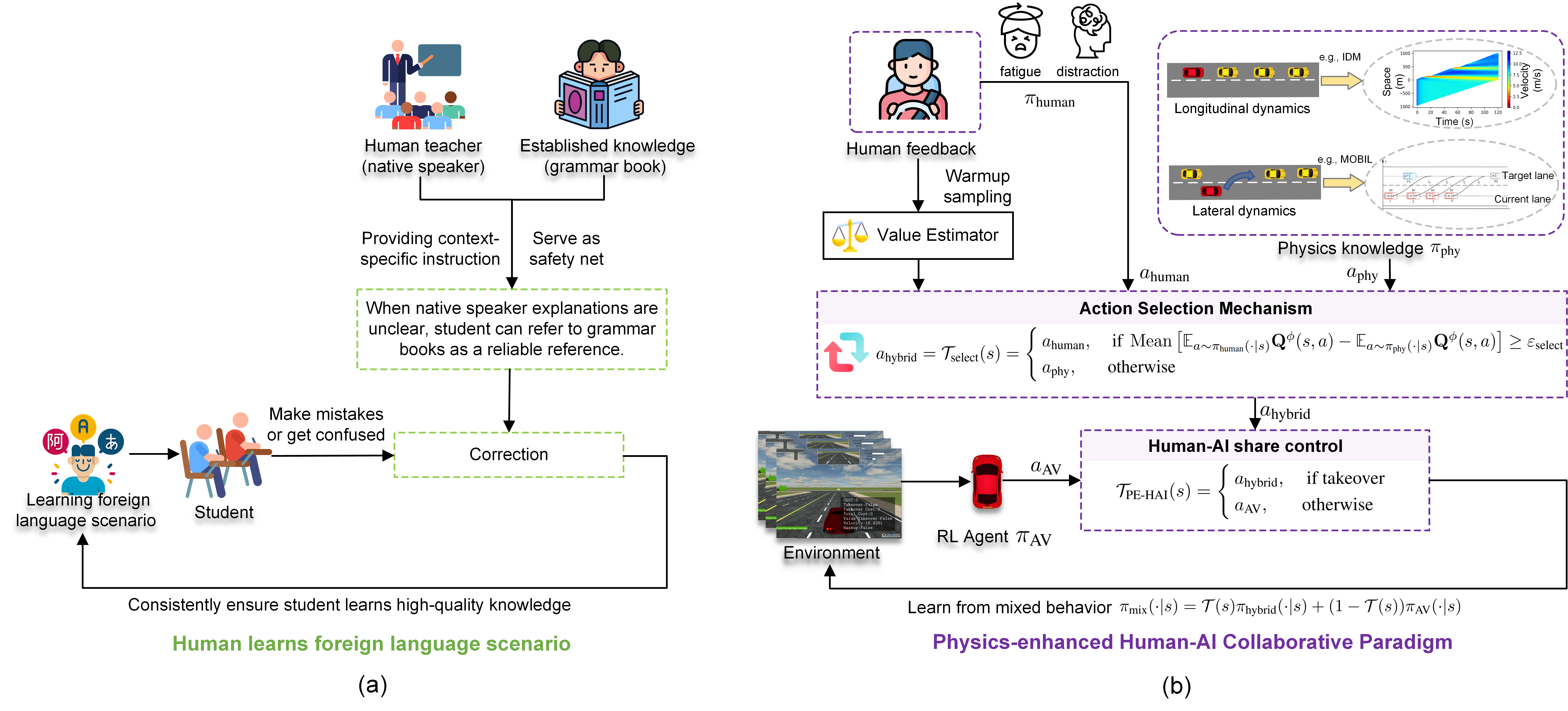}
  \caption{The proposed PE-HAI paradigm for autonomous driving. (a) Inspiration from the human learning process, where a student learns from both a native speaker and a grammar book. (b) The main components of the PE-HAI paradigm, where the AV is equipped with a human policy $\pi_{\text{human}}$, a physics-based policy $\pi_{\text{phy}}$, and an AV policy $\pi_{\text{AV}}$. When human takeover occurs, the selection function $\mathcal{T}{\text{select}}(s)$ determines whether to execute the action generated by the $\pi_{\text{human}}$ or the $\pi_{\text{phy}}$ based on their expected Q values estimated by the ensemble of Q-networks $\mathbf{Q}^\phi$.}
  \label{fig2}
\end{figure*}

As we mentioned before, most human-AI collaborative paradigms usually rely on the assumption of perfect human mentors, which may not always hold in practice. Observing the process of human learning skills,  they rely not only on human teachers but also on established knowledge. For example, as shown in Fig. \ref{fig2} (a), when learning a foreign language, a student may be guided by two mentors: a native speaker and a grammar book. The native speaker's expertise is invaluable to the student's language acquisition, providing context-specific instruction and real-world examples. Yet, in some cases, native speakers' explanations may be unclear or incorrect, for example, in the use of colloquial expressions that deviate from standard grammatical rules. In such cases, a grammar book can serve as a reliable reference and safety net, consistently ensuring that students follow the basic rules of the language. As a result, the student's language skills are improved by learning from two mentors.

Inspired by the human learning process, we propose the PE-HAI collaboration paradigm, whose main components are shown in Fig. \ref{fig2} (b). In the PE-HAI, the AV is equipped with three policies: $\pi_{\text{human}}$ (similar to the role of a native speaker), $\pi_{\text{phy}}$ (similar to the role of a grammar book), and $\pi_{\text{AV}}$ (similar to the role of the student). In detail, the $\pi_{\text{phy}}$ generates an action $a_{\text{phy}}$ based on an interpretable physics-based model, while the $\pi_{\text{human}}$ provides an action $a_{\text{human}}$ based on human judgment and situational awareness. When there is no human takeover, the AV executes $\pi_{\text{AV}}$ and learns from exploration through interaction with the environment. When humans take over, we design an action selection mechanism to determine whether the $a_{\text{human}}$ or $a_{\text{phy}}$ should be applied to the environment. In this way, although $\pi_{\text{human}}$ may occasionally fail due to factors such as fatigue, the $\pi_{\text{phy}}$ can generate feasible and safe actions in this situation. 

\subsection{Human Policy Generation}
\label{Generation of Human Policy}

\subsubsection{Human-AI Shared Control}
In this work, as shown in Fig. \ref{fig2} (b), we employ an intimate form of human-AI shared control that integrates learning from intervention (LfI) and learning from demonstration (LfD) into a unified architecture. More details about human-AI shared control can be found in our previous work \citep{huang2024human}. Specifically, we adopt a technique termed the \emph{switching function} that enables the agent to switch between exploration and intervention dynamically \citep{peng2022safe}. The switch function $\mathcal{T}$ determines the state and timing for human takeover, allowing the human to demonstrate correct actions to guide the learning agent. 

Let $\mathcal{T}(s)=1$ denote that the human takes control and $\mathcal{T}(s)=0$ mean otherwise. The mixed behavior policy $\pi_{\text{mix}}$ can be represented as:
\begin{equation}
\pi_{\text{mix}}(\cdot|s)=\mathcal{T}(s)\pi_{\text{human}}(\cdot|s)+(1-\mathcal{T}(s))\pi_{\text{AV}}(\cdot|s)
\end{equation}

A key challenge in human-AI shared control arises from the discrepancy between the training data distribution and the actual policy distribution. Specifically, $\pi_{\text{AV}}$ is trained on samples from the mixed behavior policy $\pi_{\text{mix}}$, which may not accurately represent $\pi_{\text{AV}}$'s true action distribution. A substantial difference in state distribution between these policies, quantified as $\left\|d_{\pi_{\text{mix}}} - d_{\pi_{\text{AV}}}\right\|_1$, can result in a distributional shift, adversely affecting the training process \citep{xue2023guarded}. This issue mirrors a problem in behavioral cloning (BC), where $\pi_{\text{AV}}$ learns exclusively from $\pi_{\text{human}}$ samples, albeit without interventions. To examine this state distribution discrepancy in the context of autonomous driving, we begin with a pertinent lemma \citep{achiam2017constrained}.

\begin{lemma}
\label{lemma1}
The state distribution discrepancy between the human policy $\pi_{\text{human}}$ and the AV policy $\pi_{\text{AV}}$ is bounded by their expected policy discrepancy:
\end{lemma}
\begin{equation}
\label{eq7}
\left\|d_{\pi_{\text{human}}} - d_{\pi_{\text{AV}}}\right\|1 \leqslant \frac{\gamma}{1-\gamma} \mathbb{E}_{s \sim d_{\pi_{\text{human}}}} \left\|\pi_{\text{human}}(\cdot \mid s)- \pi_{\text{AV}}(\cdot \mid s)\right\|_1
\end{equation}

We apply the lemma to the setting of human-AI shared control and derive a bound for $\left\|d_{\pi_{\text{mix}}}- d_{\pi_{\text{AV}}}\right\|_1$.

\begin{theorem}
\label{theorem1}
For any mixed behavior policy $\pi_{\text{mix}}$ deduced by a human policy $\pi_{\text{human}}$, an AV policy $\pi_{\text{AV}}$, and a switch function $\mathcal{T}(s)$, the state distribution discrepancy between $\pi_{\text{mix}}$ and $\pi_{\text{AV}}$ is bounded by
\end{theorem}
\begin{equation}
\label{eq8}
\left\|d_{\pi_{\text{mix}}}- d_{\pi_{\text{AV}}}\right\|1 \leqslant \frac{\beta\gamma}{1-\gamma} \mathbb{E}_{s \sim d_{\pi_{\text{mix}}}} \left\|\pi_{\text{human}}(\cdot \mid s)- \pi_{\text{AV}}(\cdot \mid s)\right\|_1
\end{equation}
where $\beta=\frac{\mathbb{E}_{s \sim d_{\pi_{\text{mix}}}}\left[\mathcal{T}(s)\left\|\pi_{\text{human}}(\cdot \mid s)-\pi_{\text{AV}}(\cdot \mid s)\right\|_{1}\right]}{\mathbb{E}_{s \sim d_{\pi_{\text{mix}}}}\left\|\pi_{\text{human}}(\cdot \mid s)-\pi_{\text{AV}}(\cdot \mid s)\right\|_{1}}\in[0,1]$ is the expected intervention rate weighted by the policy discrepancy.

A closer analysis of Eqs. \ref{eq7} and \ref{eq8} reveal that while both bound the state distribution discrepancy by per-state policy differences, the intervention-based upper bound is further tightened by the intervention rate $\beta$. Minimizing $\beta$ in practice can reduce this discrepancy, potentially mitigating performance degradation during deployment. Extending Thm. \ref{theorem1}, we demonstrate in \ref{Appendix A} that within the human-AI shared control setting, a similar relationship exists between the accumulated returns of $J(\pi_{\text{mix}})$ and $J(\pi_{\text{AV}})$. This finding establishes a connection between the performance of the $\pi_{\text{mix}}$ during training and the performance of the $\pi_{\text{AV}}$ during deployment.

\subsubsection{The Form of Switch Function}
In general, there are two common forms of switch function: probability-based switch function and action-based switch function \citep{peng2022safe}. In this work, we use the action-based switch function, which triggers intervention when the agent's action deviates from the human's action, such as takeover. The action-based switch function $\mathcal{T}_{\text{act}}$  is designed as \citep{huang2024human}:
\begin{equation}
\label{eq10}
\begin{aligned}
\mathcal{T}_{\text{act}}(s_t, a_t, \pi_{\text{human}}) = 
\begin{cases}
\left(a_t^{\text{human}} \sim \pi_{\text{human}}(\cdot \mid s_t), 1\right), & \text {if takeover } \\
(a_t^{\text{AV}}, 0), & \text { otherwise }
\end{cases}
\end{aligned}
\end{equation}

A Boolean indicator $I(s_t)$ denotes human takeover, and the action applied to the environment is defined as $a_t^{\text{mix}} = I(s_t) a_t^{\text{human}}+(1-I(s_t)) a_t^{\text{AV}}$. This setup eliminates unnecessary states and mitigates the safety concerns associated with traditional RL methods.

To measure the effectiveness of the $\mathcal{T}_{\text{act}}$ in the setting of the PE-HAI, we examine the return of the mixed behavior policy $J(\pi_{\text{mix}})$. With $\mathcal{T}_{\text{act}}(s)$ defined in Eq.~\ref{eq10}, $J(\pi_{\text{mix}})$ can be bounded by the following theorem:

\begin{theorem}
\label{theorem2}
With the action-based switch function $\mathcal{T}_{\text{act}}$, the return of the mixed behavior policy $J(\pi_{\text{mix}})$ is lower and upper bounded by
\end{theorem}
\begin{equation}
\begin{aligned}
J(\pi_{\text{human}})+&\frac{\sqrt{2}(1-\beta) R_{\max }}{(1-\gamma)^{2}} \sqrt{H-\varepsilon}\geqslant J(\pi_{\text{mix}})\geqslant J(\pi_{\text{human}})-\frac{\sqrt{2}(1-\beta) R_{\max }}{(1-\gamma)^{2}} \sqrt{H-\varepsilon},
\end{aligned}
\end{equation}
where $H=\mathbb{E}_{s\sim d_{\pi_{\text{mix}}}} \mathcal{H}(\pi_{\text{human}}(\cdot|s))$ represents the average entropy of the human policy during shared control, and $\beta$ is the weighted intervention rate from Thm. \ref{theorem1}. This theorem establishes a lower bound for $J(\pi_{\text{mix}})$, comprising the return of $\pi_{\text{human}}$ and an additional term linked to its entropy. Such a result suggests that $\mathcal{T}_{\text{act}}$ effectively contributes to high-return training data by enabling the AV to learn from human demonstrations. Consequently, training $\pi_{\text{AV}}$ on trajectories from $\pi_{\text{mix}}$ effectively optimizes an upper bound on the AV's suboptimality. A comprehensive proof of this theorem is provided in \ref{Appendix B}.

While our above analysis provides insights into the feasibility and efficiency of the human-AI shared control, it is crucial to note that its success is intrinsically tied to the quality of human mentors. Specifically, the bounds derived in Thm. \ref{theorem1} and Thm. \ref{theorem2} are directly related to the performance of $\pi_{\text{human}}$. Consequently,  if the performance of $\pi_{\text{human}}$ deteriorates - for instance, due to human fatigue or distraction - the method's effectiveness could degrade significantly, potentially leading to failure. 

\subsection{Physics-based Policy Generation}

Besides leveraging human feedback, we also incorporate physics knowledge into the PE-HAI to establish a trustworthy lower bound on the framework's performance. The  $\pi_{\text{phy}}$, derived from a well-established traffic flow model, serves as a reliable safeguard even in situations where human input quality may deteriorate or be inconsistent. Consistent with \cite{yang2023towards,cao2021confidence,cao2022trustworthy}, we use the intelligent driver model (IDM) \citep{treiber2000congested} and the minimizing overall braking induced by lane changes (MOBIL) model \citep{kesting2007general} to generate the action $a_{\text{phy}}$. Note that other traffic flow models may also be effective, which we leave for future research to explore and validate.  The IDM describes the longitudinal dynamics of the vehicle as follows:
\begin{equation}
\dot{v}^{\text{AV}}=\alpha\left[1-\left(\frac{v^{\text{AV}}}{{v_0}^{\text{AV}}}\right)^\eta-\left(\frac{{s^*}^{\text{AV}}(v^{\text{AV}}, {\Delta v}^{\text{AV}})}{s^{\text{AV}}}\right)^2\right]
\end{equation}
where $\dot{v}^{\text{AV}}$ and $v^{\text{AV}}$ denote the acceleration and velocity of the vehicle, respectively. $\alpha$ represents the maximum acceleration, ${v_0}^{\text{AV}}$ is the desired velocity in free-flowing traffic, $\eta$ is the exponent for velocity, ${\Delta v}^{\text{AV}}$ is the velocity difference between the vehicle and its leading vehicle, ${s^*}^{\text{AV}}$ is the desired minimum gap, and $s^{\text{AV}}$ is the actual gap.

The desired minimum gap ${s^*}^{\text{AV}}$ is determined by:
\begin{equation}
{s^*}^{\text{AV}}(v^{\text{AV}}, {\Delta v}^{\text{AV}})=s_0+\max \left(0, v T+\frac{v \Delta v}{2 \sqrt{a_{max} \beta}}\right)
\end{equation}
where $s_0$ is the standstill distance, $T$ is the safe time headway, $a_{max}$ is the maximum acceleration, and $\beta$ is the comfortable braking deceleration.

For lane-changing decisions, the MOBIL model provides an incentive criterion:
\begin{equation}
\underbrace{\tilde{a}^{\text{AV}}-a^{\text{AV}}}_{\text {AV agent}}+p^{\text{pol}}(\underbrace{\tilde{a}^{\text{new}}-a^{\text{new}}}_{\text {new follower }}+\underbrace{\tilde{a}^{\text{old}}-a^{\text{old}}}_{\text {old follower }})>\Delta a_{\text{threshold}}    
\end{equation}
where $\tilde{a}^{\text{AV}}$ and $a^{\text{AV}}$ are the anticipated and current accelerations of the vehicle, respectively. $p^{\text{pol}}$ is the politeness factor, $\tilde{a}^{\text{new}}$ and $a^{\text{new}}$ are the predicted and current accelerations of the new follower, $\tilde{a}^{\text{old}}$ and $a^{\text{old}}$ are the predicted and current accelerations of the old follower, and $\Delta a_{\text{threshold}}$ is the acceleration threshold.

\subsection{Action Selection Mechanism}

To effectively leverage the strengths of both human feedback and physics knowledge,  we design an action selection mechanism as the core module of the PE-HAI.  This mechanism, as illustrated in Fig. \ref{fig2} (b), serves as an arbitration component that evaluates and selects between actions generated by the $\pi_{\text{human}}$ or $\pi_{\text{phy}}$. Technically, we expect that the agent will choose the action with the higher expected Q value between $a_{\text{human}}$ and $a_{\text{phy}}$ to execute.

\subsubsection{Value Estimator Construction}

First, to obtain the expected Q value for $a_{\text{human}}$ and $a_{\text{phy}}$, we define two ways of constructing the value estimators: (a) Human demonstration warmup. During the warmup phase, human mentors continuously control the agent and provide high-quality action demonstrations. The human mentor's demonstration data is then used to train an estimator Q-network $Q^{\pi_{\text{est}}}$ from scratch. (b) Expert policy warmup. Following the approach in \citet{peng2022safe}, we first train an expert policy $\pi_{\text{expert}}$ in a more constrained environment. During the warmup phase, we roll out $\pi_{\text{expert}}$ and collect training samples. The collected data is then used to train the estimator Q-network $Q^{\pi_{\text{est}}}$. 

With limited training data, the estimator Q-network may fail to provide accurate estimates when encountering previously unseen states. To address this issue, we propose to use an ensemble of estimator Q-networks technique, inspired by the work of \citet{chen2021randomized}. A set of ensembled estimator Q-networks $\mathbf{Q}^{\phi}$ with the same architecture but different initialization weights are constructed and trained using the same data. 

The loss function for training the ensemble estimator $\mathbf{Q}^{\phi}$ is:
\begin{equation}
    L\left(\phi     \right)=\mathbb{E}_{s, a \sim \mathcal{D}}\left[y-\mathrm{Mean}\left[\mathbf{Q}^\phi\left(s, a\right)\right]\right]^{2},
\end{equation}
where $y=\mathbb{E}_{s^{\prime} \sim \mathcal{D},a'\sim\pi_{\text{expert}}(\cdot|s')+\mathcal{N}(0, \sigma)}\left[r+\gamma \mathrm{Mean}\left[\mathbf{Q}^\phi\left(s^{\prime}, a^{\prime}\right)\right] \right]$ is the target value, and $\mathcal{D}$ is the replay buffer storing the collected transitions $(s, a, r, s')$. By using an ensemble of estimator Q-networks, we can reduce the bias and variance in expected Q value estimates, leading to more robust and accurate value estimation.

\subsubsection{Selection Function Design}

Then, we design a selection function $\mathcal{T}_{\text{select}}(s)$ to execute the action with the higher expected Q value between $a_{\text{human}}$ and $a_{\text{phy}}$ into the environment. The $\mathcal{T}_{\text{select}}(s)$ can be formalized as follows:
\begin{equation}
\label{eq17}
   a_{\text{hybrid}} = \mathcal{T}_{\text{select}}(s) = 
    \begin{cases}
        a_{\text{human}}, & \text { if } \operatorname{Mean}\left[\mathbb{E}_{a \sim \pi_{\text{human}}(\cdot \mid s)} \mathbf{Q}^\phi(s, a)-\mathbb{E}_{a \sim \pi_{\text{phy}}(\cdot \mid s)} \mathbf{Q}^\phi(s, a)\right] \geq \varepsilon_{\text{select}} \\
        a_{\text{phy}}, & \text{otherwise}
    \end{cases}
\end{equation}
where $\varepsilon_{\text{select}}$ is a small threshold to account for estimation errors. $a_{\text{hybrid}} $ represents the hybrid intervention action generated by the selection function with the higher expected return among the $\pi_{\text{human}}$ and $\pi_{\text{phy}}$. 

Now, combining Eqs. \ref{eq10} and \ref{eq17}, we can define the final mathematical expression of the PE-HAI collaborative paradigm as: 
\begin{equation}
\label{eq18}
    \mathcal{T}_{\text{PE-HAI}}(s) = 
    \begin{cases}
        a_{\text{hybrid}}, & \text { if takeover} \\
        a_{\text{AV}}, & \text{otherwise}
    \end{cases}
\end{equation}

When there is no takeover, the AV agent executes $a_{\text{AV}}$ and learns from exploration through interaction with the environment. When a takeover occurs, it determines whether the $a_{\text{human}}$ or $a_{\text{phy}}$ should be applied to the environment. In detail, when the $a_{\text{human}}$ has a higher expected Q value than the $a_{\text{phy}}$, the PE-HAI trusts the human's judgment and selects $a_{\text{human}}$. Otherwise, the $a_{\text{phy}}$ will be executed to maintain the safety lower bound. This allows the PE-HAI to generate trustworthy and safe actions, even when the human mentor occasionally makes a suboptimal decision.

\subsubsection{Analysis of Trustworthy Safety Improvement}

To demonstrate the safety improvement of the PE-HAI, we analyze its performance in comparison to using either $\pi_{\text{human}}$ and $\pi_{\text{phy}}$ alone. As mentioned above, the goal is to learn an optimal policy $\pi_{\text{AV}}^*$. Analyzing Eq. \ref{eq4}, we find that the higher the quality of human feedback in RLHF, the closer the learned $\pi_{\text{AV}}$ is to $\pi_{\text{AV}}^*$. In the setting of PE-HAI, combining Eqs. \ref{eq1} , \ref{eq17}, and \ref{eq18}, we can obtain the satisfaction of the constraints in Eq. \ref{eq19}.

\begin{theorem}
\label{theorem3}
The expected cumulative reward obtained by learning from $\pi_{\text{hybrid}}$ is equal to the maximum of the expected cumulative rewards obtained by $\pi_{\text{human}}$ and  $\pi_{\text{phy}}$. It is also guaranteed to be greater than or equal to the expected cumulative reward obtained by the $\pi_{\text{phy}}$. 
\end{theorem}
\begin{equation}
\label{eq19}
\begin{aligned}
\mathbb{E}_{\tau \sim \pi_{\text{hybrid}}}\left[\sum_{t=0}^{H} \gamma^t r(s_t, a_t)\right] & = \max\left(\mathbb{E}_{\tau \sim \pi_{\text{human}}}\left[\sum_{t=0}^{H} \gamma^t r(s_t, a_t)\right], \mathbb{E}_{\tau \sim \pi_{\text{phy}}}\left[\sum_{t=0}^{H} \gamma^t r(s_t, a_t)\right]\right) \\
& \geq \mathbb{E}_{\tau \sim \pi_{\text{phy}}}\left[\sum_{t=0}^{H} \gamma^t r(s_t, a_t)\right]
\end{aligned}
\end{equation}

\begin{figure*}[t]
\centering
  \includegraphics[width=0.6\textwidth]{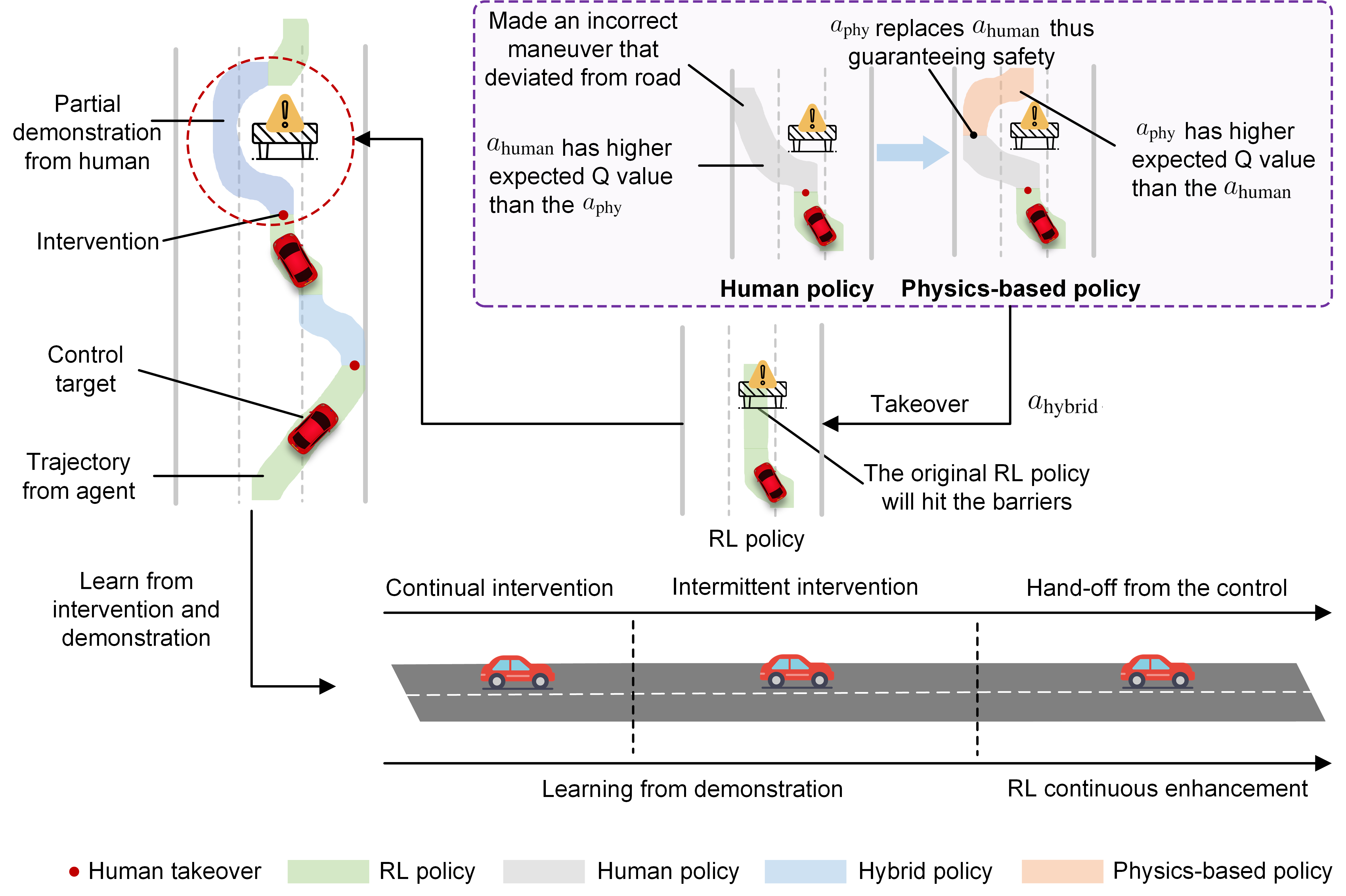}
  \caption{PE-HAI's action selection process in a roadblock avoidance scenario. Traditional RL methods (green path) would collide with the roadblock before learning to avoid it. In PE-HAI, humans perceive danger and take over, making a left lane change. As $a_{\text{human}}$'s expected Q value exceeds $a_{\text{phy}}$ at this point, PE-HAI adopts the human action (gray path). However, humans may subsequently make erroneous maneuvers such as deviating from the road. While this could cause training failure in traditional RLHF, PE-HAI switches to the physics-based action when $a_{\text{phy}}$'s Q value surpasses $a_{\text{human}}$'s as the vehicle nears road departure (orange path), thus ensuring training safety. Ultimately, the agent learns a safe and efficient obstacle avoidance strategy from this hybrid policy (blue path).}
  \label{fig3}
\end{figure*}

The detailed derivation of Thm. \ref{theorem3} is provided in \ref{Appendix C}. According to Thm. 1 in \cite{cao2022trustworthy}, the expected cumulative reward of a policy can serve as an objective measure for evaluating driving safety. Therefore, Eq. \ref{eq19} indicates that the PE-RLHF framework can ensure superior driving safety performance since it learns from $\pi_{\text{hybrid}}$. In other words, even when human mentors occasionally make suboptimal decisions, the PE-RLHF framework can still guarantee that the safety performance is at least as good as the existing interpretable $\pi_{\text{phy}}$.  Fig. \ref{fig3}. illustrates the advantages of PE-HAI in a roadblock avoidance scenario. The overall workflow of the PE-HAI is shown in \ref{Appendix:Workflow-PE-HAI} as pseudocode.

\section{Physics-enhanced Reinforcement Learning with Human Framework}
\label{Physics-enhanced Reinforcement Learning with Human Framework}

In this section, we propose a PE-RLHF framework, as shown in Fig. \ref{fig4}. The whole framework consists of five parts: (a) Observation space and action space, (b) Reward-free actor-critic architecture, (c) Learning from hybrid intervention action, (d) Learning from exploration with entropy regularization, and (e) Reducing the human mentor's cognitive load. In the following subsections, we will explain each of these components in detail.

\subsection{Observation Space and Action Space }

\begin{figure*}[t]
\centering
  \includegraphics[width=0.9\textwidth]{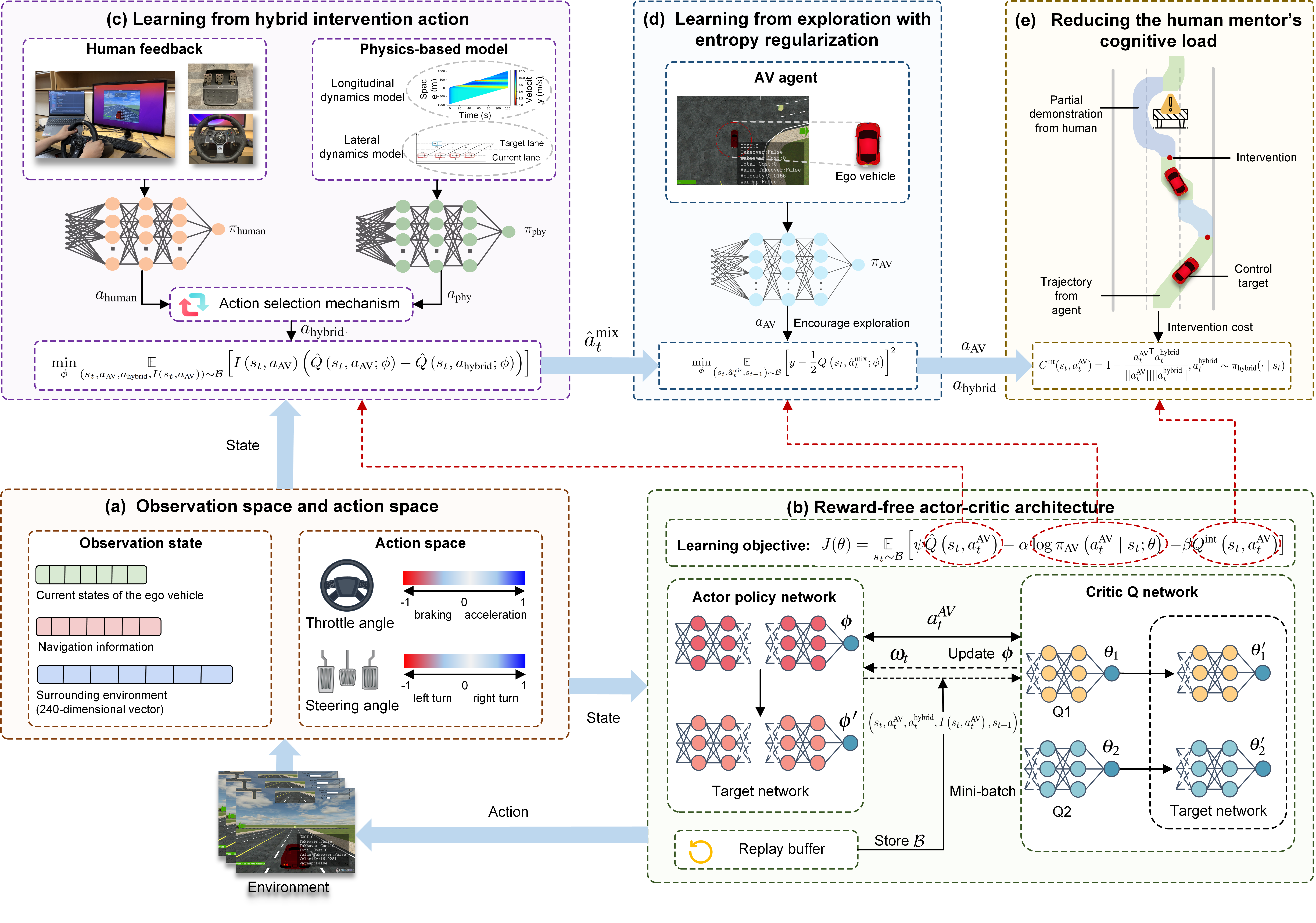}
  \caption{Overview of the proposed PE-RLHF framework.}
  \label{fig4}
\end{figure*}

Following the end-to-end learning paradigm, we design observation space and action space to directly map raw sensory inputs (LiDAR) to control commands (throttle and steering angle), minimizing the need for intermediate representations. As shown in Fig. \ref{fig4} (a), the observation state consists of three parts, designed to provide a comprehensive view of the driving environment:  (a) Ego state includes current states of the ego vehicle, such as steering angle, heading angle, velocity, and relative distance to road boundaries. (b) Navigation information includes the relative positions of the target vehicle concerning the upcoming checkpoints. (c) The surrounding environment uses a 240-dimensional vector to represent the 2D-Lidar-like point clouds, capturing the surrounding environment within a maximum detecting distance of 50m, centered at the target vehicle. Each entry in this vector is normalized to the range [0, 1], indicating the relative distance of the nearest obstacle in the specified direction.  

Different from methods that pre-select a subset of actions as candidates \citep{cao2022trustworthy}, we employ a more challenging approach by defining the action space as a continuous space bounded by [-1, 1]. This continuous action space design allows for smoother and more precise control, enabling the agent to learn more nuanced driving behaviors. Specifically, the action is defined as the throttle and steering angle. For steering wheel control, negative values represent left turn commands, while positive values correspond to right turn commands. Regarding the throttle, negative values indicate braking commands and positive values correspond to acceleration commands.

\subsection{Removing the Reward Function}
\label{Reward-free}

Some RLHF-enabled works try to reshape a reward function from human demonstration data to avoid manual reward design \citep{wu2022prioritized,wang2024reinforcement, zhuang2024hgrl}. Nevertheless, this method still faces challenges such as potential bias in offline demonstration data and difficulty in capturing complex human preferences. Upon reevaluating our primary objective, we realize that a conventional reward function is not necessary. Instead, our core aim is to incorporate human preferences into the learning process. Human intervention serves as a clear indicator of the agent's suboptimal performance, whether due to safety concerns or inadequate behavior. Conversely, the absence of intervention implies that the agent's actions align with human expectations. This binary feedback mechanism effectively encodes human preferences without the need for a traditional reward structure. 

Drawing from this insight, we propose replacing the traditional reward function with a proxy value function to represent human preferences. This approach eschews explicit rewards, focusing instead on calculating proxy Q values \citep{huang2024human,li2022efficient}. The key advantage lies in our ability to manipulate these proxy Q values to elicit desired behaviors, leveraging the value-maximizing nature of value-based RL, as demonstrated in Eq. \ref{eq1}. By omitting immediate rewards, we transform the standard Q value update $Q^{\pi}(s_t, a_t) \leftarrow R(s, a) + \gamma \max_{a_{t+1}} Q^{\pi}(s_{t+1}, a_{t+1})$ into a proxy Q value update $\hat{Q}^{\pi}(s_t, a_t) \leftarrow \gamma \max_{a_{t+1}} \hat{Q}^{\pi}(s_{t+1}, a_{t+1})$. Subsequently, we can derive the policy $\pi_{\theta}$ by optimizing
\begin{equation}
\label{eq20}
\theta = \arg\max_{\theta} \mathbb{E}_{(s,a) \sim \rho^{\pi}} \hat{Q}^{\pi}(s,a)
\end{equation}

While Eq. \ref{eq20} employs proxy Q values, it maintains the MDP structure without tracking explicit rewards. The temporal difference (TD)-based approach initially reassigns proxy Q values to partial human demonstrations before propagating these values across states. The policy is then optimized to align with human intentions as captured by the proxy value function. Section \ref{Learning from Hybrid Intervention Action} details the implementation, while \ref{Appendix D} offers a convergence theorem and proof validating the proxy value function's efficacy. Eq. \ref{eq20} enables us to reframe standard RL into a reward-free paradigm that learns from active human engagement. This approach circumvents the challenging task of manual reward function design, a particularly complex undertaking in domains such as autonomous driving.

\subsection{Learning Objectives for Value Network}

We propose a comprehensive set of objectives that can effectively utilize human feedback and physics knowledge. The learning objectives are as follows: (a) The agent should aim to maximize the proxy value function, denoted as $\hat{Q}(s, a)$, which reflects the value of the hybrid intervention action $a_{\text{hybrid}} $. (b) The agent should actively explore the state-action space. This is achieved by maximizing the entropy of the action distribution, denoted as $\mathcal{H}(\pi(\cdot \mid s))$. (c) The agent should strive to reduce the cognitive load of the human mentor by minimizing the intervention value function, denoted as $Q^{\text{int}}(s, a)$. 

The overall learning objective can be summarized as follows:
\begin{equation}
\max_\pi \mathbb{E}\left[\hat{Q}(s, a)+\mathcal{H}(\pi)-Q^{\text{int}}(s, a)\right]
\label{Eq15}
\end{equation}

In the following sections, we will delve into the practical implementation details of each design objective mentioned above. 

\subsubsection{Learning from Hybrid Intervention Action}
\label{Learning from Hybrid Intervention Action}
According to the observation in Section~\ref{Reward-free}, we should strive to make the agent's behavior close to the behavior selected by the PE-HAI, which combines human and physics knowledge. A closer analysis of Eq. \ref{eq1} indicates that the optimal deterministic strategy consistently selects the action with the highest Q value. Consequently, in states where human intervention occurs, $a_{\text{hybrid}}$ should invariably have higher values than alternative actions, while agent actions $a_{\text{AV}}$ should have comparatively lower values.

The mixed behavior policy $\pi_{\operatorname{\text{mix}}}(a \mid s)$ generates transition sequences $\left\{\left(s_t, a_t^{\text{AV}},a_t^{\text{hybrid}}, I\left(s_t,  a_t^{\text{AV}}\right), s_{t+1}\right), \ldots\right\}$, which serve as partial demonstrations. These, along with free exploration transitions, are stored in a replay buffer $\mathcal{B}$ and integrated into the training pipeline, without recording environmental rewards or costs. Learning solely from these partial demonstrations in $\mathcal{B}$ introduces a distribution shift. To mitigate this, we employ CQL \citep{kumar2020conservative} for off-policy training. We sample $\left(s_t, a_t^{\text{AV}},a_t^{\text{hybrid}}, I\left(s_t, a_t^{\text{AV}}\right)\right)$ from $\mathcal{B}$, assigning proxy Q values $\hat{Q}\left(s_t, a^{\text{hybrid}}\right)$ to $a_t^{\text{hybrid}}$ and $\hat{Q}\left(s_t, a^{\text{AV}}\right)$ to $a_t^{\text{AV}}$. The optimization problem of the proxy value function is formulated as follows:
\begin{equation}
\min_\phi \underset{\left(s_t, a_{\text{AV}}, a_{\text{hybrid}}, I\left(s_t, a_{\text{AV}}\right)\right) \sim \mathcal{B}}{\mathbb{E}} \left[I\left(s_t, a_{\text{AV}}\right)\left(\hat{Q}\left(s_t, a_{\text{AV}} ; \phi\right) -\hat{Q}\left(s_t, a_{\text{hybrid}} ; \phi\right)\right)\right]
\label{eq22}
\end{equation}

This optimization objective embodies an optimistic bias towards $a_{\text{hybrid}}$ while maintaining a pessimistic outlook on the agent's action $a_{\text{AV}}$. By minimizing the proxy Q value discrepancy between $a_{\text{hybrid}}$ and $a_{\text{AV}}$, as expressed in Eq. \ref{eq22}, we effectively steer the agent's behavior towards the high-value state-action subspace favored by PE-HAI.

\subsubsection{Learning from Exploration with Entropy Regularization}

Insufficient exploration of the PE-HAI's preferred subspace during free sampling can result in rare encounters with high proxy value states. This scarcity hinders the backward propagation of proxy values, potentially impeding the learning process. To address this issue and promote more comprehensive exploration, we incorporate entropy regularization \citep{haarnoja2018soft}, which introduces an auxiliary signal for proxy value function updates:

\begin{equation}
\min_\phi \underset{\left(s_t, \hat{a}_t^{\text{mix}}, s_{t+1}\right) \sim \mathcal{B}}{\mathbb{E}}\left[y-\frac{1}{2}Q\left(s_t, \hat{a}_t^{\text{mix}} ; \phi\right)\right]^2
\label{eq23}
\end{equation}
where
\begin{equation}
y = \gamma \underset{a_{t+1} \sim \pi_{\text{AV}}\left(\cdot \mid s_{t+1}\right)}{\mathbb{E}}\left[Q\left(s_{t+1}, a_{t+1} ; \phi^{\prime}\right)-\alpha \log \pi_{\text{AV}}\left(a_{t+1} \mid s_{t+1}\right)\right]
\label{eq24}
\end{equation}
where $\hat{a}_t^{\text{mix}}$ is the action performed under state $s_t$, $\phi^{\prime}$ represents the delayed update parameters of the target network, and $\gamma$ is the discount factor.

As the PE-RLHF framework operates in a reward-free setting, we omit the reward term from the update target $y$. Combining Eqs.~\ref{eq23} and \ref{eq24}, we formulate the comprehensive optimization objective for the proxy value function as:
\begin{equation}
\min_\phi \underset{\mathcal{B}}{\mathbb{E}}\left[\left(y-\frac{1}{2}Q\left(s_t, \hat{a}_t^{\text{mix}} ; \phi\right)\right)^2 + I\left(s_t, a_t^{\text{AV}}\right)\left(\hat{Q}\left(s_t, a_t^{\text{AV}} ; \phi\right)-\hat{Q}\left(s_t, a_t^{\text{hybrid}} ; \phi\right)\right)\right]
\label{eq25}
\end{equation}

Unlike traditional offline RL approaches that rely on static datasets without closed-loop feedback \citep{kumar2020conservative}, PE-RLHF leverages both online exploration and partial hybrid action data. Moreover, it maintains continuity in state visitation between human mentor and agent, thereby effectively addressing potential distribution shift concerns.

\subsubsection{Reducing the Human Mentor’s Cognitive Load}

Unrestricted PE-HAI intervention frequency may lead to the agent's over-reliance on $a_{\text{hybrid}}$, potentially compromising its performance when evaluated independently~\citep{peng2022safe,li2022efficient,wu2023toward}. This vulnerability stems from $\hat{Q}(s_t,a_t)$ reflecting the proxy Q value of $\pi_{\text{mix}}$ rather than $\pi_{\text{AV}}$. Consequently, the agent might choose actions contradicting PE-HAI preferences, such as boundary violations, necessitating frequent interventions. This cycle perpetuates low automation and imposes a high cognitive burden on the human mentor due to constant corrective action requirements.

To reduce the human mentor's cognitive load and increase the AV's autonomy, we introduce a subtle penalty for agent behaviors that prompt PE-HAI intervention. This penalty is quantified using the cosine similarity between $a_{\text{AV}}$ and $a_{\text{hybrid}}$, serving as an intervention cost~\citep{li2022efficient}. The formulation is as follows:

\begin{equation}
C^{\text{int}}(s_t, a_t^{\text{AV}}) = 1-\cfrac{{a_t^{\text{AV}}}^{\mathsf{T}}a_t^{\text{hybrid}}}{||a_t^{\text{AV}}|| ||a_t^{\text{hybrid}}||}, a_t^{\text{hybrid}} \sim \pi_{\text{hybrid}}(\cdot \mid s_t)
\label{eq26}
\end{equation}

The agent faces substantial penalty only when $a_{\text{AV}}$ and $a_{\text{hybrid}}$ exhibit significant cosine dissimilarity. ~\cite{li2022efficient} demonstrated this method's superiority over fixed penalties like a '+1' cost. Additionally, we attribute the intervention cost to the agent only during the initial step of intervention by the PE-HAI. This approach is grounded in the observation that intervention by the PE-HAI, triggered by a specific action $a_t^{\text{AV}}$, signifies a deviation from the preferences of the PE-HAI at that particular moment.

By reducing the occurrence of such actions, the level of automation of the agent can be increased, thus reducing the cognitive load of human mentor. To mitigate the abuse of intervention by the PE-HAI, we introduce an intervention value function, denoted as $Q^{\text{int}}(s, a_t^{\text{AV}})$, which represents the expected cumulative cost of intervention by the PE-HAI. This method parallels the technique of estimating state-action values via the Bellman equation in Q-learning.
\begin{equation}
Q^{\text{int}}(s_t, a_t^{\text{AV}}) = C^{\text{int}}(s_t,a_t^{\text{AV}}) + \gamma \mathbb{E}{s_{{t+1}\sim \mathcal B, a_{t+1}\sim \pi_{\text{AV}}(\cdot | s_{t+1})}}[Q^{\text{int}}(s_{t+1},a_{t+1}^{AV})]
\label{eq27}
\end{equation}

The value function is employed to optimize the policy directly.

\subsection{Learning Policy for Policy Network}

The policy network is responsible for determining control actions and strives to optimize the value network. The batch gradient of the policy network can be expressed as follows:

\begin{equation}
\begin{aligned}
J(\theta) = \underset{s_t \sim \mathcal{B}}{\mathbb{E}} \left[\psi \hat Q\left(s_t, a_t^{\text{AV}}\right)-\alpha \log \pi_{\text{AV}}\left(a_t^{\text{AV}} \mid s_t ; \theta\right) \right. \left. - \beta Q^{\text{int}}\left(s_t, a_t^{\text{AV}}\right)  \right]
\end{aligned}
\label{eq28}
\end{equation}
where the entropy regularization coefficient $\alpha$ enhances the policy by encouraging a balance between exploitation and exploration. The coefficient $\psi$ weights the importance of the proxy Q value, reflecting the emphasis on aligning the agent's actions with the learned value function. Meanwhile, $\beta$ serves as a weighting factor for the intervention value function, allowing for a controlled trade-off between the agent's autonomy and its reliance on intervention by the PE-HAI. The overall workflow of the PE-RLHF is shown in \ref{Appendix:Workflow-PE-RLHF} as pseudocode.

\section{Experimental Evaluation}
\label{Experimental Evaluation}

In this section, we will conduct experiments to investigate the following questions for evaluating our proposed PE-RLHF method: (a) Can PE-RLHF learn driving policies with higher learning efficiency and performance compared to other methods that do not consider human feedback and physics knowledge? (b) Can PE-RLHF provide safety guarantees and achieve trustworthy performance improvement compared to other RLHF methods that do not leverage physics knowledge, especially when the quality of human feedback deteriorates over time? (c) Is PE-RLHF robust under different traffic environments and human feedback quality?

To answer questions (a) and (b), we compare PE-RLHF with physics-based methods, RL and safe RL methods, offline RL and IL methods, and RLHF methods. For question (c), we compare different proficiency levels of human mentors, as well as different traffic environments and parameter settings.

\subsection{Experiment Setup}
\subsubsection{Experiment Environment}
Considering the potential risks associated with involving human subjects in physical experiments, we benchmarked the different methods in a lightweight driving simulator MetaDrive~\citep{li2022metadrive}, which retains the ability to evaluate safety and generalizability in unseen environments. MetaDrive employs procedural generation techniques to synthesize an infinite number of driving maps, enabling the separation of training and testing sets, which facilitates benchmarking the generalization capabilities of various methods in the context of safe driving. The simulator is also extremely efficient and flexible, allowing us to run the human-AI collaboration experiment in real time. The player's goal is to drive the ego vehicle to a pre-determined destination, avoiding dangerous behaviors such as collisions.

\begin{figure*}[t]
\centering
  \includegraphics[width=0.99\textwidth]{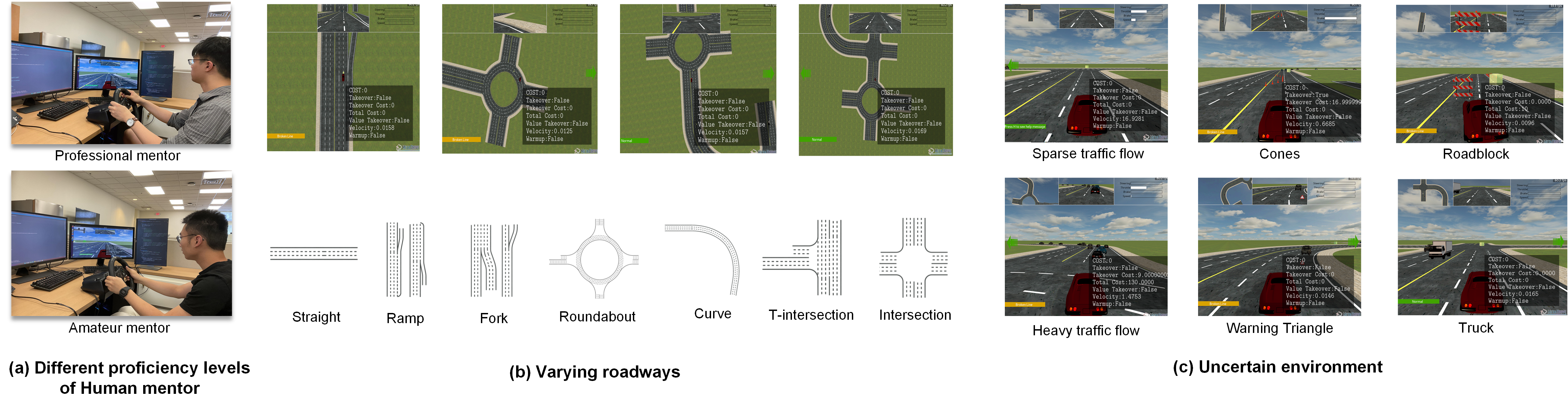}
  \caption{Illustration of various driving scenarios generated in the MetaDrive simulator.}
  \label{fig5}
\end{figure*}

\subsubsection{Scenario Description}
To validate the performance of PE-RLHF in more realistic traffic environments, we utilize the MetaDrive simulator to generate traffic scenarios with varying roadways (e.g., straight, ramp, fork, roundabout, curve, T-intersection, and intersections). Fig. \ref{fig5} illustrates some of the generated driving scenarios in MetaDrive. The driving environment is designed to be full of uncertainty, similar to natural driving environments. The designed uncertain environment has the following features: 

(a) \textbf{Surrounding Vehicle:} Each surrounding vehicle has an unobservable behavior generation model. The behavior generation model uses MetaDrive's default rule-based planner control. Each surrounding vehicle generates an action at each time step based on the situation of the surrounding vehicles. Therefore, each surrounding vehicle has a unique driver behavior when interacting with other vehicles. The ego vehicle must consider surrounding vehicles' normal driving behaviors, such as following lanes, changing lanes, and exiting roundabouts. This setup simulates the strong uncertainty present in real-world driving. 

(b) \textbf{Random Traffic:} The generation time and location of surrounding vehicles are random. These vehicles influence each other. When the ego vehicle starts driving in the simulator, the simulator first randomly generates surrounding vehicles the ego vehicle. Additionally, the surrounding vehicles are of various types, including trucks and cars. The combination of different types of surrounding vehicles and their random initial positions form increasingly complex traffic scenarios.

(c) \textbf{Random Obstacles:} We randomly generate obstacles, such as stationary broken-down vehicles, stationary traffic cones, and triangular warning signs. The driving road is set to three lanes, allowing the ego vehicle to change lanes to avoid collisions. Collisions can occur in various ways. Such an environment presents a high-dimensional driving problem, making it difficult to design a perfect policy that handles all situations. However, this setting is more realistic for natural driving, as real driving processes are not simple concatenations of individual cases

The uncertain environments cause the ego vehicle to encounter different surrounding traffic, forcing the generated driving policies to adapt to the uncertain surroundings, addressing the over-fitting issue.

\subsubsection{Reward and Cost Definition}
Although PE-RLHF does not rely on environmental rewards during the training phase, we provide a reward function for training the baseline method and evaluating different methods during testing. Specifically, we
follow the default reward scheme in MetaDrive. The reward function is composed of three parts as follows \citep{li2022metadrive}:
\begin{equation}
R=c_1 R_{\text {driving}}+c_2 R_{\text {speed}}+R_{\text {destination }}
\end{equation}
where the driving reward $R_{\text {driving }}=d_t-d_{t-1}$, wherein the $d_t$ and $d_{t-1}$ denote the longitudinal coordinates of the target vehicle on the current reference lane of two consecutive time steps, providing a dense reward to encourage the agent to move toward the destination. The speed reward $R_{\text {speed }}=v_t / v_{\max }$ incentivizes agent to drive fast. $v_t$ and $v_{\max }$ denote the current velocity and the maximum velocity (80 km/h), respectively. The termination reward $R_{\text {termination }}$ contains a set of sparse rewards. At the end of the episode, other dense rewards will be disabled and only one sparse reward will be given to the agent at the end of the episode according to its termination state. We set $c_1=1$, $c_2=0.1$, and $R_{\text {destination }}=20$. 

In addition, if the ego vehicle collisions with vehicles, obstacles, sidewalks, and buildings, a `+1' is added to the cost at each time step. Note that PE-RLHF does not have access to this cost during training.

\subsection{Baselines}
To benchmark the proposed PE-RLHF for autonomous driving, we conduct experimental comparisons with state-of-the-art methods. We categorize them into several groups: Physics-based methods, RL and Safe RL methods, Offline RL and IL methods, and RLHF methods.

\begin{itemize}
\item \textbf{Physics-based Methods.} These methods rely solely on predefined physics-based models to generate driving actions, without any learning or human feedback. Consistent with \cite{cao2022trustworthy,tang2022highway}, we use a combination of the IDM \citep{treiber2000congested} for longitudinal control and the MOBIL \citep{kesting2007general} for lateral control. Both are widely used driving models, but other driving models can also be employed.
\item \textbf{RL and Safe RL Methods.} SAC-RS \citep{tang2022highway} and PPO-RS \citep{ye2020automated} use reward shaping (RS) technique to address the issue of potentially diminished learning efficiency when the reward signals generated by the environment are sparse. SAC-Lag~\citep{ha2021learning}, PPO-Lag~\citep{stooke2020responsive}, and CPO~\citep{achiam2017constrained} aim to improve safety during the RL training process by imposing constraints on policy optimization.
\item \textbf{Offline RL and IL Methods.} CQL~\citep{kumar2020conservative} learns from a fixed dataset collected by human mentors without access to online exploration. It addresses the distribution shift problem in offline RL by learning a conservative Q-function that lower bounds the true Q-function. BC~\citep{sharma2018behavioral} and GAIL~\citep{kuefler2017imitating} learn from human demonstrations to mimic expert behavior. BC directly learns a policy that maps states to actions, while GAIL learns a reward function that encourages the agent to behave similarly to the expert.
\item \textbf{RLHF Methods.} HG-DAgger~\citep{kelly2019hg} and IWR~\citep{mandlekar2020human} integrate human intervention data into the training buffer and perform behavior cloning to update the policy. HG-DAgger allows human mentors to intervene during the agent's exploration, while IWR re-weights the intervention data based on the frequency of human takeover. HACO~\citep{li2022efficient} and HAMI-DRL~\citep{huang2024human} allow human mentors and AI agents to share autonomy in the training process, aiming to improve safety and efficiency.
\end{itemize}

By comparing PE-RLHF with these methods, we can demonstrate how PE-RLHF improves the learning efficiency, driving performance, and safety of training by leveraging both human feedback and physics knowledge. 

\subsection{Evaluation Strategy}
\subsubsection{Evaluation Metric}
To comprehensively assess the performance of the PE-RLHF and compare it with other approaches, we introduce a set of metrics that capture various aspects of autonomous driving performance: (a) \textbf{Episodic Return.} The cumulative reward obtained by the agent in an episode. (b) \textbf{Success Rate.} The percentage of episodes where the agent reaches the destination while staying within the road boundaries. (c) \textbf{Safety Violation.} The total cost incurred due to collisions with vehicles or obstacles in an episode. (d) \textbf{Travel Distance.} The distance covered by the agent in each episode. (e) \textbf{Travel Velocity.} The average velocity maintained by the agent during an episode. (f) \textbf{Total Overtake Count.} The number of vehicles overtaken by the agent in each episode.

\subsubsection{Three-stage Strategy}
During the evaluation, the metrics mentioned above should be combined and sequentially inspected, as they assess the driving performance from different aspects. For instance, a high overtake count might suggest a good agent, but this is only true if the agent also maintains a decent success rate, stays within the road boundaries, and keeps the safety violation reasonably low. 

In this work, we propose a three-stage strategy to evaluate the methods' performance: (a) \textbf{Stage I.} In the first stage, we focus on the \textit{episodic return} and \textit{success rate}. These metrics provide a high-level assessment of the agent's overall driving performance and its reliability in completing the task. (b) \textbf{Stage II.} If the methods demonstrate similar performance in Stage I, we proceed to the second stage, which examines the \textit{safety violation} and \textit{travel distance}. These metrics provide a more detailed evaluation of the agent's ability to avoid collisions and cover distance safely. (c) \textbf{Stage III.} In cases where the methods exhibit comparable performance in both Stage I and Stage II, we move to the third stage, which analyzes the \textit{travel velocity} and \textit{total overtake count}. These metrics assess the agent's ability to navigate efficiently in a traffic environment. The three-stage evaluation strategy allows for a hierarchical assessment of the methods' performance and helps identify the best-performing agents in a structured manner.

\subsection{Implementation Details}
We conduct experiments using RLLib, a distributed learning system that allows large-scale parallel experiments. All experiments were conducted on a high-performance desktop computer running Ubuntu 20.04, equipped with an Intel Core i9-10980XE CPU, two Nvidia GeForce RTX 4090 GPUs, and 128GB RAM. Each trial consumes 2 CPUs with 8 parallel rollout workers. Additionally, our tests indicate that the PE-RLHF can also run successfully on a lower-configuration computer equipped with a Nvidia GeForce RTX 2080 Ti GPU. The physics-based methods were repeated three times, while the RL, secure RL, offline RL, and IL experiments were repeated five times using different random seeds. All RLHF experiments were repeated three times except for HG-DAgger and IWR. Owing to restricted human resources, both the ablation studies and sensitivity analysis experiments are conducted only once. 

In traditional RL, agents are typically trained and evaluated in the same fixed environment, which can lead to over-fitting and poor performance in unseen scenarios. To evaluate the generalization performance of different methods, we split the driving scenes into a training set and a test set, each containing 50 distinct scenes. After each training iteration, we assess the learning agent's performance in the test environments using the same seed without human intervention and record the evaluation metrics. Information about other hyper-parameters is given in \ref{Appendix:Hyper-parameters}.

\begin{table*}[!t]
\centering
\begin{small}
\caption{The performance of different baselines in the MetaDrive simulator.}
\label{tab1}
\renewcommand{\arraystretch}{2} 
\setlength{\tabcolsep}{1.5pt}
\resizebox{\textwidth}{!}{%
\begin{tabular}{@{}cc|cccc|cc|cc|cc@{}}
\toprule
\multirow{3}{*}{\textbf{Category}} & \multirow{3}{*}{\textbf{Method}} & \multicolumn{4}{c|}{\textbf{Training}} & \multicolumn{6}{c}{\textbf{Testing}} \\ \cline{7-12} & &  & & & & \multicolumn{2}{c|}{Stage I} & \multicolumn{2}{c|}{Stage II} & \multicolumn{2}{c}{Stage III} \\
\cmidrule(lr){3-6} \cmidrule(lr){7-12}
& & \multirow{1}{*}{\shortstack{Human Data\\Usage}} \multirow{1}{*}{$\downarrow$} & \multirow{1}{*}{\shortstack{Total Data\\Usage}} \multirow{1}{*}{$\downarrow$} & \multirow{1}{*}{\shortstack{Training\\Time}} \multirow{1}{*}{$\downarrow$} & \multirow{1}{*}{\shortstack{Total Safety\\Violation}} \multirow{1}{*}{$\downarrow$} & \multirow{1}{*}{\shortstack{Episodic\\Return}} \multirow{1}{*}{$\uparrow$}& \multirow{1}{*}{\shortstack{Success Rate \\ (\%)}} \multirow{1}{*}{$\uparrow$} & \multirow{1}{*}{\shortstack{Safety \\ Violation}} \multirow{1}{*}{$\downarrow$} & \multirow{1}{*}{\shortstack{Travel \\ Distance}} \multirow{1}{*}{$\uparrow$} & \multirow{1}{*}{\shortstack{Travel \\ Velocity}} \multirow{1}{*}{$\uparrow$} & \multirow{1}{*}{\shortstack{Total Overtake \\Count}} \multirow{1}{*}{$\uparrow$}\\
\toprule
Expert & Human Demo. & 49K & - & - & -  & 388.16 {\tiny $\pm$45.00} & 1  & 0.03 {\tiny $\pm$0.00} & - & - & 0 \\
\midrule
\shortstack{Physics-based} & IDM-MOBIL & - & - & - & -& 206.30 {\tiny $\pm$35.23}& 0.31 {\tiny $\pm$0.15}  & 0.49 {\tiny $\pm$0.08} & 108.56 {\tiny $\pm$55.23} & 19.78 {\tiny $\pm$2.67} & 0 {\tiny $\pm$0}\\
\midrule
\multirow{5}{*}{\shortstack{RL and\\Safe RL}} & SAC-RS & - & 1M & 38h & 1.13K {\tiny $\pm$0.38K}  & 305.01 {\tiny $\pm$21.84} & 0.65 {\tiny $\pm$0.13} & 0.52 {\tiny $\pm$0.19} & 134.37 {\tiny $\pm$12.04} & 27.85 {\tiny $\pm$ 4.69} & 0 {\tiny $\pm$0}\\
& PPO-RS & - & 1M & 30h & 0.59K {\tiny $\pm$0.10K}  & 276.52 {\tiny $\pm$72.53} & 0.47 {\tiny $\pm$0.24} & 2.15 {\tiny $\pm$0.34} & 123.11{\tiny $\pm$34.00} & \textbf{30.58} {\tiny $\pm$2.91} & 0 {\tiny $\pm$0}\\
& SAC-Lag & - & 1M & 39h & 1.18K {\tiny $\pm$0.50K}  & 297.20 {\tiny $\pm$17.65} & 0.58 {\tiny $\pm$0.09} &  0.51 {\tiny $\pm$0.10} & 143.42 {\tiny $\pm$9.59} & 26.77{\tiny $\pm$3.66} & 0 {\tiny $\pm$0}\\
& PPO-Lag & - & 1M & 32h & 0.42K {\tiny $\pm$0.18K}  & 232.70 {\tiny $\pm$50.99} & 0.28 {\tiny $\pm$0.09} & 1.61 {\tiny $\pm$0.64} & 115.25{\tiny $\pm$26.48} & 25.68{\tiny $\pm$2.02} & 0 {\tiny $\pm$0}\\
& CPO & - & 1M & - & 4.36K {\tiny $\pm$2.22K} & 194.06 {\tiny $\pm$108.86} & 0.21 {\tiny $\pm$0.29}  & 1.71 {\tiny $\pm$1.02}  & - & - & -\\ 
\midrule
\multirow{3}{*}{\shortstack{Offline RL\\and IL}} & CQL & 49K (1.0) & - & 92h & 1257.27{\tiny $\pm$ 119.41}  & 81.07 {\tiny $\pm$11.20} & 0.01 {\tiny $\pm$0.02} & 1.11 {\tiny $\pm$0.10} & 29.62 {\tiny $\pm$5.65} & 20.89 {\tiny $\pm$0.26} & 0 {\tiny $\pm$0}\\
& BC & 49K (1.0) & -  & 94h & 131.45{\tiny $\pm$ 30.56}  & 0.01 {\tiny $\pm$0.01} & 0 {\tiny $\pm$0} & \textbf{0.20} {\tiny $\pm$0.00} & 8.03 {\tiny $\pm$1.88} & 0.52 {\tiny $\pm$0.32} & 0 {\tiny $\pm$0}\\
& GAIL & 49K (0.20) & - & 120h & 766.68 {\tiny $\pm$6.48} & 0 {\tiny $\pm$0} & 0 {\tiny $\pm$0} &  1.05{\tiny $\pm$0.12} & 1.50 {\tiny $\pm$0.06} & 6.64 {\tiny $\pm$0.48c} & 0 {\tiny $\pm$0}\\
\midrule
\multirow{4}{*}{\shortstack{RLHF}} & HG-DAgger & 38.52K (0.77) & 50K  & 3h & 52.49  & 93.25 & 0.20 & 1.43 & 48.46 & 16.00 & 0\\
& IWR & 35.48K (0.71) & 50K  & 3h & 48.47  & 227.73 & 0.61 & 1.64 & 113.96 & 20.37 & 0\\
& HACO &  9.21K (0.31) & 30K  & 1h &  36.59 {\tiny $\pm$11.64} & 340.73 {\tiny $\pm$10.55} & 0.82 {\tiny $\pm$0.06} &  1.45 {\tiny $\pm$0.98} & 174.28 {\tiny $\pm$14.08}  &  19.22 {\tiny $\pm$3.30} & 7.67 {\tiny $\pm$1.89}\\
& HAIM-DRL \textsuperscript{*} & 8.22K (0.27) & 30K & 1h  & 29.84 {\tiny $\pm$ 10.25}  & 354.34 {\tiny $\pm$ 11.08 } & 0.85 {\tiny $\pm$0.03} &  0.76 {\tiny $\pm$0.28} & - & - & -\\
\midrule
\multirow{1}{*}{\shortstack{Physics-enhanced \\ RLHF}} & \textbf{PE-RLHF (Ours)} \textsuperscript{*} & \textbf{7.86K (0.26)} & \textbf{30K}  & \textbf{1h} & \textbf{16.61} {\tiny $\pm$9.96} & \textbf{391.48} {\tiny $\pm$20.47} & \textbf{0.85} {\tiny $\pm$0.04} & 0.47 {\tiny $\pm$0.01} & \textbf{177.00} {\tiny $\pm$3.74} & 21.85 {\tiny $\pm$0.02} & \textbf{16.33} {\tiny $\pm$4.61}\\

\bottomrule
\end{tabular}%
}
\end{small}
\begin{flushleft}
\textsuperscript{*} In this study, we relaxed the assumption of perfect human mentors to consider more realistic conditions. Different from HAIM-DRL, which allowed only one rigorous experimental error, we permitted up to five errors in all RLHF experiments. For PE-RLHF, we default to using human demonstration warmup. For CQL, we use a dataset size of 50K transitions. The overtake count represents the total number of overtakes across all episodes. 
\end{flushleft} 
\end{table*}

\subsection{Performance Comparison}
\begin{figure*}[!t]
\centering
  \includegraphics[width=0.85\textwidth]{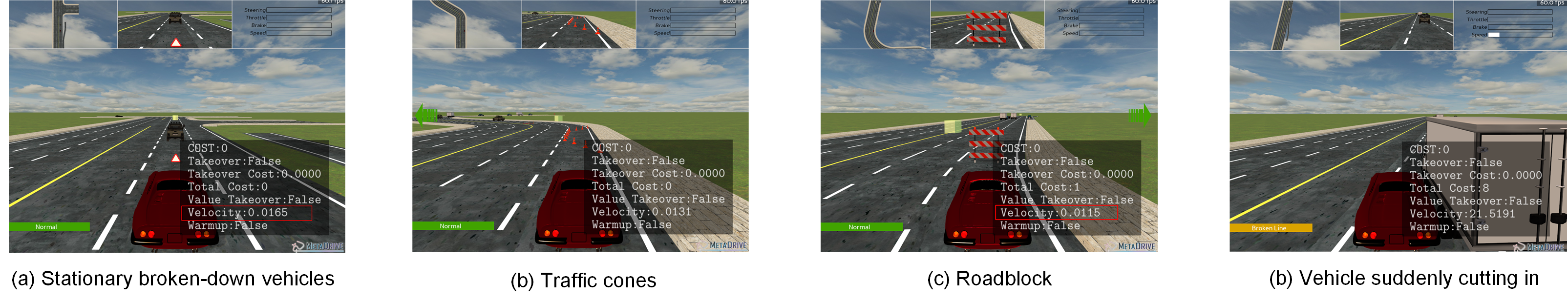}
  \caption{Typical failure scenarios of physics-based methods in complex driving environments. (a) Stationary broken-down vehicles. (b) Traffic cones. (c) Roadblock. (d) Sudden lane-cutting vehicles.}
  \label{fig6}
\end{figure*}

\begin{figure*}[!t]
  \centering
  \includegraphics[width=0.85\textwidth]{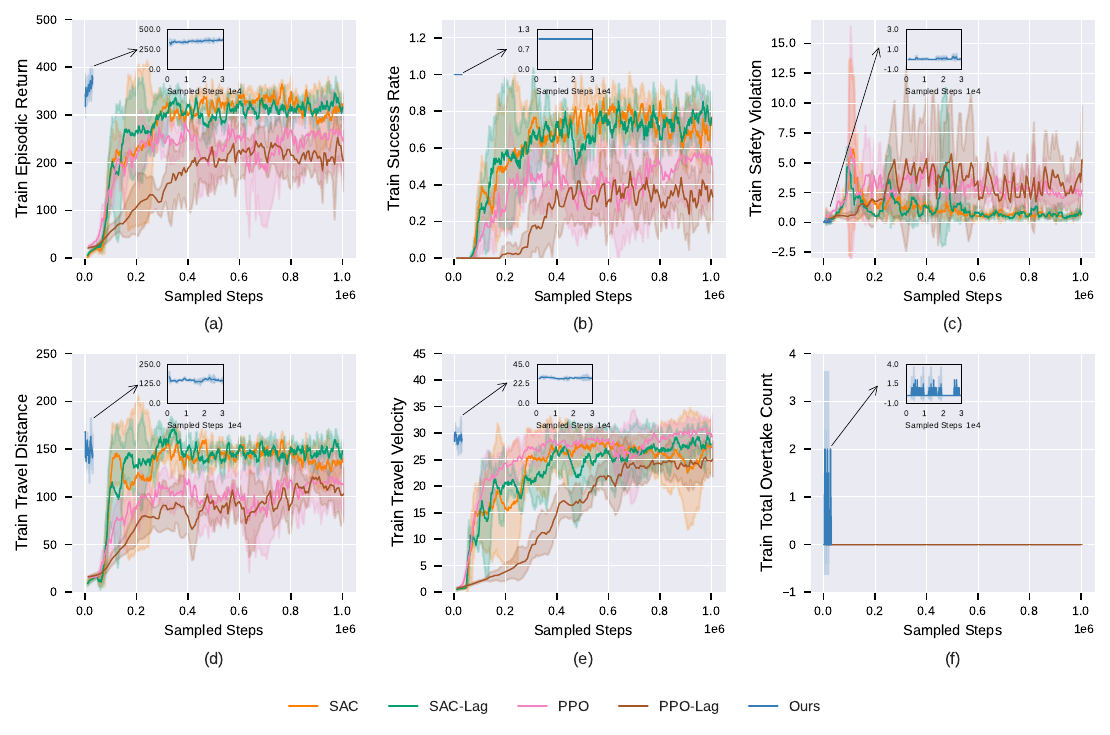}
  \caption{Performance comparison of PE-RLHF with conventional RL and Safe RL methods during the training phase.}
  \label{fig7}
\end{figure*}

\begin{figure*}[!t]
  \centering
  \includegraphics[width=0.85\textwidth]{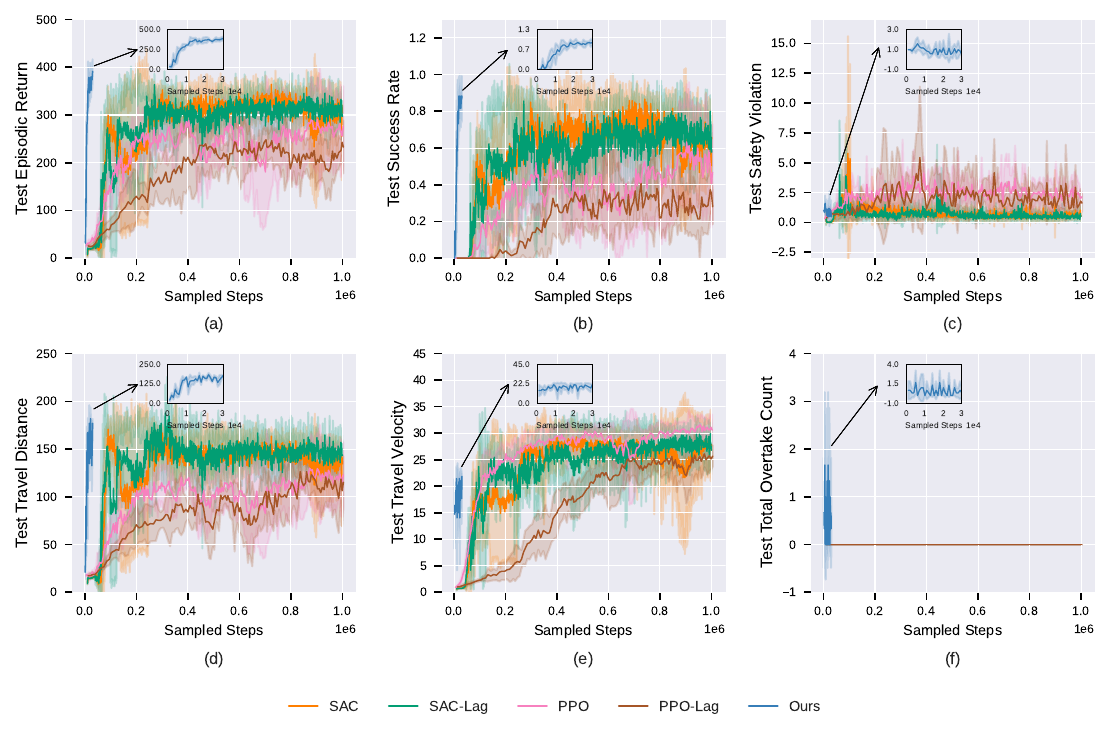}
  \caption{Performance comparison of PE-RLHF with conventional RL and Safe RL methods during the testing phase.}
  \label{fig8}
\end{figure*}

The compared results are summarized in Tab. \ref{tab1} and Figs. \ref{fig6} - \ref{fig13}. Tab. \ref{tab1} represents the average of the maximum values observed at the last checkpoint of each evaluation. Bold numbers represent the best-performing metrics for each corresponding measure. Data for physics-based methods are averaged over multiple runs. In all figures, the solid line represents the average value across different random seeds, and the shaded area indicates the standard deviation.

\subsubsection{Compared to Physics-based Methods}

We first compare PE-RLHF with the physics-based methods. The parameters of the IDM and MOBIL models are shown in \ref{Appendix:Hyper-parameters}. To ensure safe driving, all parameters are set slightly more conservatively. For example, the desired minimum gap and time headway in IDM are set larger so that the safety distance can be maintained. From Tab. \ref{tab1}, we can observe significant improvements across various metrics. In Stage I, PE-RLHF demonstrates superior overall performance with an episodic return of 391.48, far exceeding IDM-MOBIL's 206.30. Additionally, the success rate of PE-RLHF (0.85) far surpasses that of IDM-MOBIL (0.31). These results show that PE-RLHF is more effective in completing driving tasks and reaching destinations. In Stage II, PE-RLHF exhibits enhanced safety performance with a lower safety violation and a greater travel distance.  Compared to RL or Safe RL methods, IDM-MOBIL can guarantee lower safety violations. Nevertheless, we found that due to its strict rules, it tends to be too conservative in complex driving scenarios and cannot execute efficient operations such as overtaking. 

We found that IDM-MOBIL's lower success rate is because physics-based methods mainly consider interactions between vehicles and struggle to effectively handle situations with fixed obstacles, even if these obstacles can be detected by sensors such as LiDAR. Fig. \ref{fig6} illustrates several typical scenarios where physics-based methods often become stuck when faced with stationary broken-down vehicles, traffic cones, and roadblocks, failing to take measures such as lane changes. Furthermore, when vehicles from adjacent lanes suddenly cut in, physics-based methods often fail to avoid obstacles in time, thereby increasing the risk of collision. This is mainly because physics-based models are designed primarily for ideal traffic flow situations and struggle to cope with sudden and non-standard traffic events. In contrast, the proposed PE-RLHF  can better handle these complex situations by learning from human driver feedback and experience. Overall, the PE-RLHF agent can significantly improve the performance of given physics-based driving policies, which has important practical implications for autonomous driving.

\nopagebreak

\subsubsection{Compared to RL and Safe RL Methods}
To evaluate the performance of PE-RLHF against RL and Safe RL methods, we compare it with SAC-RS, PPO-RS, SAC-Lag, PPO-Lag, and CPO. The results are summarized in  Tab. \ref{tab1}  and illustrated in Figs. \ref{fig7}  and \ref{fig8} for both training and testing phases. Examining the training process, we observe that PE-RLHF exhibits remarkable safety performance. Throughout the entire training phase, it recorded only 16.61 safety violations. Despite not explicitly considering environmental costs, this result represents a reduction of two orders of magnitude compared to other RL methods. For instance, SAC-Lag and SAC-RS recorded 1.18K and 1.13K safety violations, respectively. 

During the testing, in Stage I, PE-RLHF demonstrates superior performance in terms of episodic return and success rate. The episodic return of PE-RLHF (391.48) significantly outperforms all RL and Safe RL methods, with the closest competitor being SAC-RS (305.01). Similarly, PE-RLHF achieves the highest success rate (0.85), substantially surpassing other methods, with SAC-RS again being the nearest competitor (0.65). These results indicate that PE-RLHF can complete driving tasks more effectively and reach destinations more reliably than other methods. \cite{zhou2023accelerating} reported that PPO and SAC performed poorly on the safe driving task in the MetaDrive, even in the dense reward setting. Therefore, it is not surprising that the performance is also poor under the more difficult sparse reward condition used in this work. Due to its underwhelming safety performance, CPO's results are omitted from Figs. \ref{fig7}  and \ref{fig8} to maintain the focus on more successful methods. 

Moving to Stage II, PE-RLHF continues to excel. It exhibits the lowest safety violation (0.47) among all methods, indicating superior safety performance. The next best performer in this metric is SAC-Lag (0.51). PE-RLHF also achieves the highest travel distance (177.00m), surpassing SAC-Lag (143.42m), which suggests that PE-RLHF can navigate longer distances safely. This combination of low safety violations and high travel distance implies that PE-RLHF can maintain safety over extended periods of driving. In Stage III, PE-RLHF maintains its dominance. It achieves the highest total overtake count (16.33), outperforming other methods. An interesting phenomenon observed is that all RL and safe RL methods have no overtaking records. We observed that this is because these methods tend to train the ego vehicle to be quite conservative, often waiting for all surrounding vehicles to pass before starting to move. This behavior, while safe, can lead to inefficient driving and potentially cause traffic congestion in real-world scenarios. The results highlight the advantages of integrating human feedback and physics knowledge in RL for autonomous driving tasks.

\subsubsection{Compared to Offline RL and IL Methods}

To evaluate offline RL and IL methods, we first collected a human demonstration dataset comprising almost one hour of human demonstrations. This dataset contains approximately 49,000 transitions in the training environment \footnote{The high-quality demonstration dataset collected by the human mentor in this study is available at: \href{https://github.com/zilin-huang/PE-RLHF/releases/tag/v1.0.0}{\textcolor{magenta}{https://github.com/zilin-huang/PE-RLHF/releases/tag/v1.0.0}}}. This high-quality demonstration dataset achieves 100\% success rate, with an episodic return of 388.16 and a low safety violation rate of 0.03, thus establishing a benchmark for our evaluation. By leveraging this dataset, we compare PE-RLHF with CQL, BC, and GAIL. To ensure a comprehensive evaluation, we introduced randomness by setting the spawn point to either true or false. Additionally, for CQL, we tested two variants with different iteration counts: 50k and 100k.

\begin{figure*}[!t]
  \centering
  \includegraphics[width=0.85\textwidth]{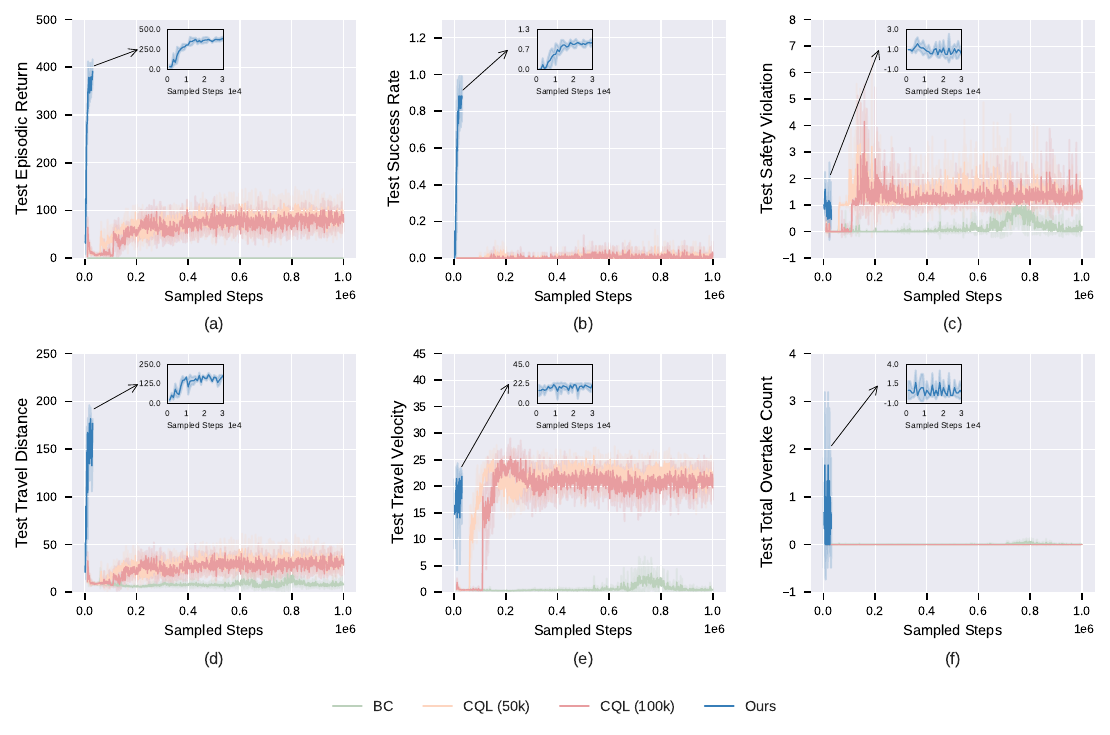}
  \caption{Performance comparison of PE-RLHF with offline RL and IL methods.}
  \label{fig9}
\end{figure*}

\begin{figure*}[!t]
  \centering
  \includegraphics[width=0.85\textwidth]{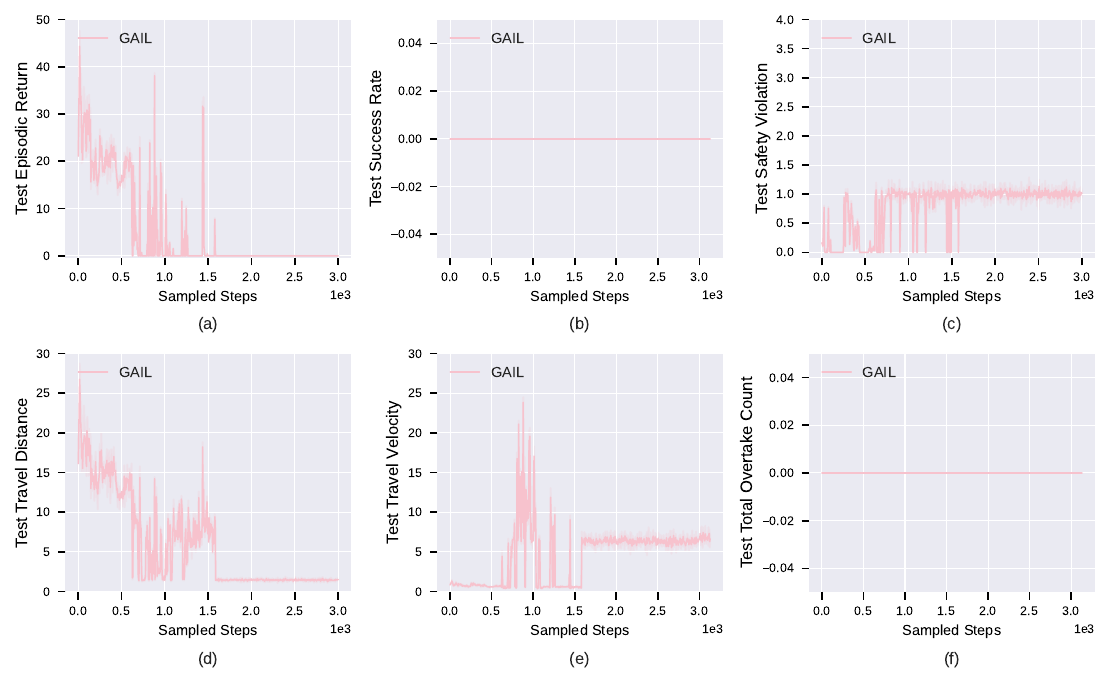}
  \caption{Performance comparison of PE-RLHF with GAIL.}
  \label{fig10}
\end{figure*}

Examining the results in Tab. \ref{tab1}  and Figs. \ref{fig9}  and \ref{fig10} , we observe significant differences in performance across these methods. During testing, in Stage I, PE-RLHF substantially outperforms all offline RL and IL methods in terms of episodic return and success rate. PE-RLHF achieves an episodic return of 391.48 and a success rate of 0.85, which are orders of magnitude higher than the best-performing offline method, CQL (50k), which only achieves an episodic return of 81.07 and a success rate of 0.01. BC and GAIL perform even worse, with near-zero episodic returns and success rates. In Stage II,  the seemingly superior safety of BC and GAIL is due to the almost non-existent forward movement of the AV. This can be verified by the travel distance. 

The poor performance of offline RL and IL methods can be attributed to several factors. These methods struggle with distribution shifts, finding it challenging to generalize beyond the demonstration data, especially in the dynamic and unpredictable environment of autonomous driving. They also suffer from a lack of exploration, as different from PE-RLHF, which can interact with the environment and learn from its experiences, offline methods are limited to the fixed dataset they are trained on. Note that increasing the number of CQL iterations from 50K to 100K did not result in a significant performance gain, suggesting that the method may have reached its upper performance limit.

\nopagebreak

\subsubsection{Compared to RLHF Methods}

To compare PE-RLHF with other RLHF methods, we categorized these into two groups: those that do not utilize online exploration data (HG-DAgger and IWR) and those that do (HACO and HAIM-DRL). 

\begin{figure*}[t]
  \centering
  \includegraphics[width=0.85\textwidth]{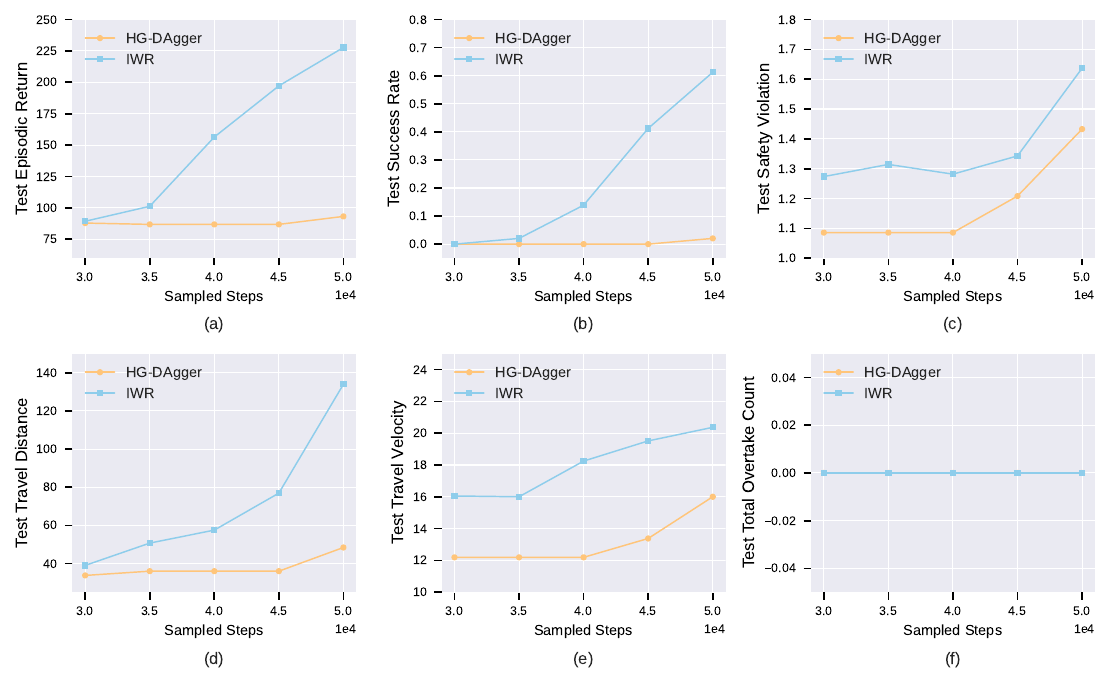}
  \caption{Performance comparison of PE-RLHF with offline RLHF methods.}
  \label{fig11}
\end{figure*}

\textit{1) Comparison with Offline RLHF Methods.} We benchmarked the performance of HG-DAgger and IWR, using the human dataset collected as described earlier. Both of these methods necessitate a preliminary warmup phase, which involves behavior cloning from the pre-collected dataset. Following the initial warmup phase, both HG-DAgger and IWR integrated human intervention data into their training buffers. Subsequently, they executed behavior cloning to update the policy over 4 epochs. Fig. \ref{fig11} demonstrated the performance changes in the warmup phase for both methods using a set of transitions containing 10 - 50K transitions. We observe that both HG-DAgger and IWR show significant improvements in performance as the warmup dataset size increases. At 10K transitions, both methods struggle to learn effective driving policies, with low episodic returns and success rates. As we increase the dataset size to 30K, we see a notable performance improvement, particularly for IWR. 

The performance continues to improve up to 50K transitions, where we see the best results for both methods. This finding is consistent with the results from \cite{li2022efficient}. Specifically, with 50K warmup transitions, in Stage I, HG-DAgger achieves an episodic return of 93.25, while IWR performs better with an episodic return of 227.73. Notably, only IWR manages to reach an acceptable success rate (0.61). This is likely because it prioritizes human intervention samples and successfully learns critical operations, avoiding dangerous situations caused by compounding errors. In contrast, HG-DAgger struggles to learn from the limited number of critical human demonstrations. Nevertheless, neither HG-DAgger nor IWR performs as well as PE-RLHF. Moving to Stage II, HG-DAgger and IWR show higher safety violations (1.43 and 1.64 respectively) and lower travel distances (48.46m and 113.96m respectively). 

\begin{figure*}[!t]
  \centering
  \includegraphics[width=0.85\textwidth]{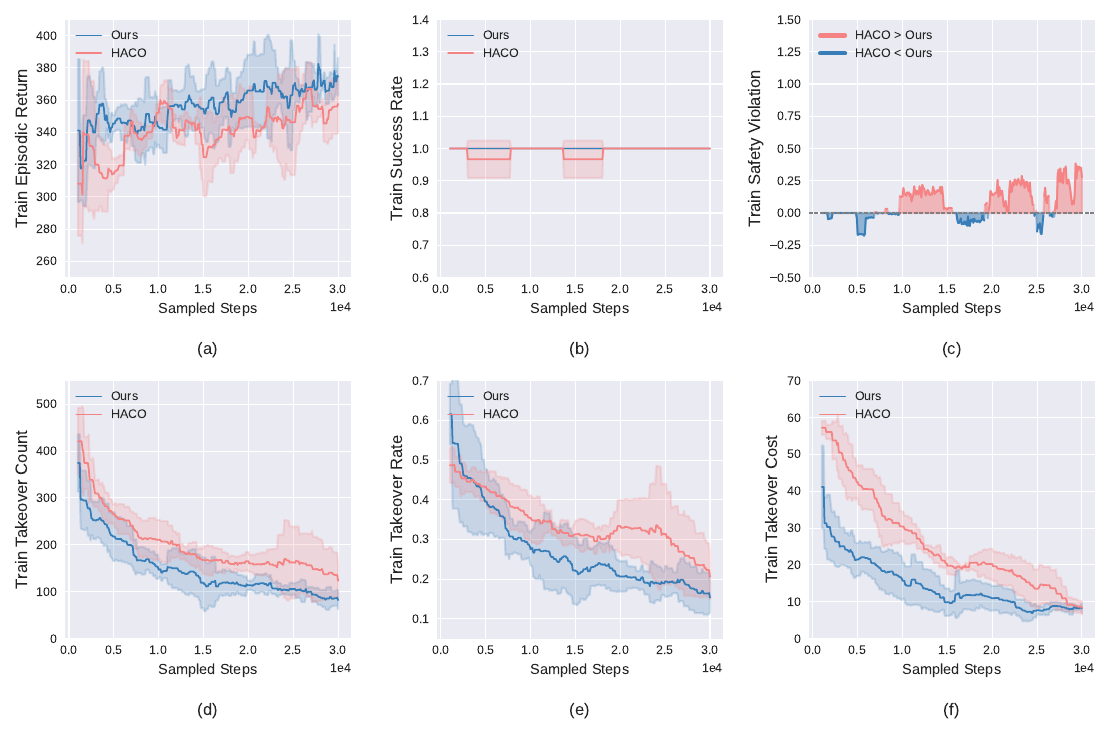}
  \caption{Performance comparison of PE-RLHF with HACO during the training phase by professional mentor.}
  \label{fig12}
\end{figure*}

\begin{figure*}[!t]
  \centering
  \includegraphics[width=0.85\textwidth]{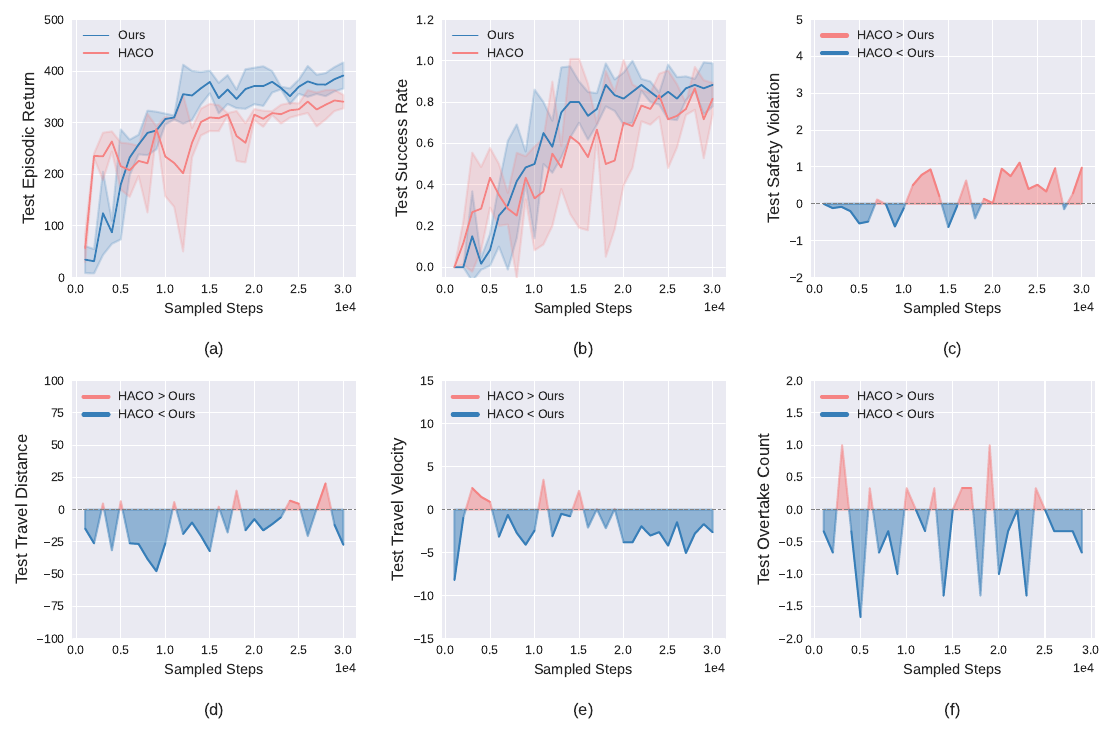}
  \caption{Performance comparison of PE-RLHF with HACO during the testing phase by professional mentor.}
  \label{fig13}
\end{figure*}

\textit{2) Comparison with Online RLHF Methods.} We compare PE-RLHF with two state-of-the-art online RLHF methods: HACO \citep{li2022efficient} and HAIM-DRL \citep{huang2024human}. The experimental data for both methods are presented in Tab. \ref{tab1}. Notably, this study places a greater emphasis on safety performance, whereas HAIM-DRL also considers the balance of traffic flow efficiency. Therefore, in Figs. \ref{fig12} and \ref{fig13}, we focus on depicting the training and testing processes for HACO and our proposed PE-RLHF method. 

A crucial observation is the significant difference in train safety violations. PE-RLHF records only 16.61 safety violations during the entire training process, compared to HACO's 36.59. The superior safety performance of PE-RLHF can be attributed to the integration of physics knowledge. The incorporation of well-established traffic models such as IDM and MOBIL provides a trustworthy lower bound for safety performance, even when the quality of human demonstration deteriorates. This is evident from  Fig. \ref{fig12} (b) and (c). Moreover, a notable observation from  Fig. \ref{fig12} (d) and (e) is that PE-RLHF requires fewer human takeovers and maintains a lower takeover rate throughout the training process compared to HACO. This suggests that PE-RLHF learns to drive safely more quickly, reducing the need for human intervention. Furthermore, as seen in Fig. \ref{fig12} (f), PE-RLHF incurs lower takeover costs, indicating that when interventions do occur, they are less severe or prolonged than those required by HACO. 

During testing, in Stage I, PE-RLHF demonstrates superior performance compared to both HACO and HAIM-DRL. PE-RLHF achieves an episodic return of 391.48 and a success rate of 0.85, outperforming HACO (340.73 and 0.82 respectively) and HAIM-DRL (354.34 and 0.85 respectively). As shown in Fig. \ref{fig13} (a) and (b), PE-RLHF consistently achieves higher episodic returns and success rates than HACO throughout the testing process. Moving to Stage II, PE-RLHF continues to excel with the lowest safety violation (0.47) compared to HACO (1.45) and HAIM-DRL (0.76).  Fig. \ref{fig13} (c)  clearly illustrates PE-RLHF's superior safety performance throughout the testing process. Regarding travel distance, PE-RLHF achieves 177.00m, slightly higher than HACO's 174.28m, as shown in  Fig. \ref{fig13} (d). In Stage III, PE-RLHF maintains its dominance with the highest travel velocity (21.85km/h) and total overtake count (16.33). Fig. \ref{fig13} (e) and (f)  show that PE-RLHF consistently achieves higher travel velocities and more frequent overtaking maneuvers compared to HACO.

\nopagebreak

\subsection{Sampling Efficiency Analysis}
\begin{figure*}[!t]
  \centering
  \includegraphics[width=0.75\textwidth]{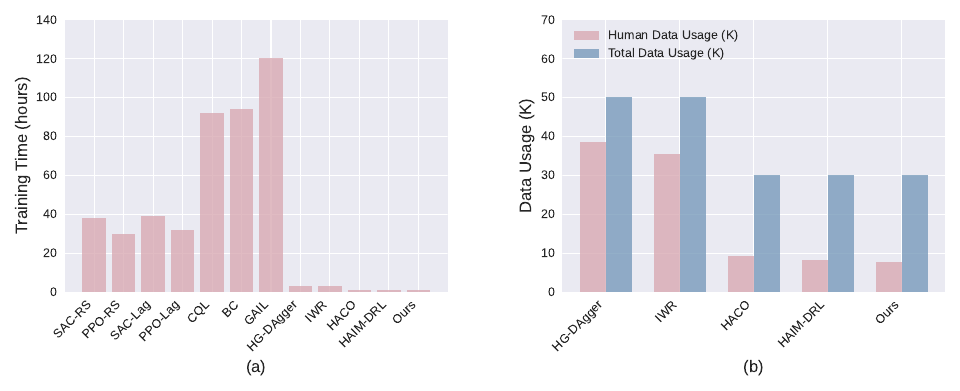}
  \caption{Sampling efficiency and computational performance comparison of PE-RLHF with other methods.}
  \label{fig14}
\end{figure*}

Compared with other baseline methods, as illustrated in Fig. \ref{fig14}, PE-RLHF demonstrates excellent sampling efficiency and computational performance. It achieves a test success rate of 0.85 with only 30K environmental interactions, of which merely 7.86K (26\%) were safe operation steps provided by human demonstrators. Compared to traditional RL and Safe RL methods, PE-RLHF reduces training time from at least 30 hours to just 1 hour - a nearly 30-fold improvement. On the other hand, CQL requires 92 hours of training with 49K  human demonstration data. BC and GAIL use similar amounts of demonstration data and require 94 and 120 hours of training time respectively.  The superior performance of PE-RLHF method can be attributed to its ability to learn directly from high-quality human demonstrations rather than relying solely on trial-and-error exploration. 

Compared with HG-DAgger and IWR, which require 80 hours of training and use 38.52K (77\%) and 35.48K (71\%) human demonstration data respectively, PE-RLHF achieves superior performance with significantly less data and time. Meanwhile, when compared to HACO and HAIM-DRL, PE-RLHF shows a slight advantage in data efficiency (26\% vs. 31\% and 27\% respectively) while maintaining comparable computational efficiency. Furthermore, PE-RLHF converges to stable performance at around 15K iterations, while HACO, for instance, requires about 25K iterations to stabilize. The superior performance achieved with less data and shorter training time underscores the effectiveness of PE-RLHF in combining human feedback, physics knowledge, and RL. Such improvements in sampling efficiency and computational performance could have significant implications for the practical implementation of autonomous driving systems, potentially reducing development cycles and resource requirements.

\subsection{Sensitivity Analysis}

\subsubsection{Impact of Physics-based Model}

\begin{table*}[t]
\centering
\begin{small}
\caption{Performance comparison of PE-RLHF with different physics-based model combinations.}
\label{tab2}
\renewcommand{\arraystretch}{2} 
\setlength{\tabcolsep}{1pt}
\resizebox{\textwidth}{!}{%
\begin{tabular}{@{}cc|c|cc|cc|cc@{}}
\toprule
\multirow{3}{*}{\textbf{Method}} & \multirow{3}{*}{\shortstack{\textbf{Driving Operation}}} & \multirow{2}{*}{\shortstack{\textbf{Training}}} & \multicolumn{6}{c}{\textbf{Testing}}  \\ \cline{4-9} & &  & \multicolumn{2}{c|}{Stage I} & \multicolumn{2}{c|}{Stage II} & \multicolumn{2}{c}{Stage III} \\ \cline{3-9}
& & \multirow{1}{*}{\shortstack{Total Safety\\Violation}} \multirow{1}{*}{$\downarrow$}  & \multirow{1}{*}{\shortstack{Episodic\\Return}} \multirow{1}{*}{$\uparrow$} & \multirow{1}{*}{\shortstack{Success Rate \\ (\%)}} \multirow{1}{*}{$\uparrow$}  & \multirow{1}{*}{\shortstack{Safety \\ Violation}} \multirow{1}{*}{$\downarrow$} & \multirow{1}{*}{\shortstack{Travel \\ Distance}}  \multirow{1}{*}{$\uparrow$} & \multirow{1}{*}{\shortstack{Travel\\Velocity}}  \multirow{1}{*}{$\uparrow$}  & \multirow{1}{*}{\shortstack{Total Overtake \\Count}}  \multirow{1}{*}{$\uparrow$} \\ \midrule
IDM-MOBIL & Longitudinal \& Lateral & - & 206.30 {\tiny $\pm$35.23}& 0.31 {\tiny $\pm$0.15} & 0.49 {\tiny $\pm$0.08} & 108.56 {\tiny $\pm$55.23} & 19.78 {\tiny $\pm$2.67} & 0 {\tiny $\pm$0} \\ \midrule
PE-RLHF (without) & - & 39.45 {\tiny $\pm$ 12.32} & 302.67 {\tiny $\pm$ 21.88} & 0.73 {\tiny $\pm$ 0.05} & 1.48 {\tiny $\pm$ 0.43} & 138.23 {\tiny $\pm$ 4.28} & 16.58 {\tiny $\pm$ 0.96} & 6.14 {\tiny $\pm$ 1.12} \\ \midrule
PE-RLHF (with IDM) & Longitudinal only & 28.79 {\tiny $\pm$ 9.97} & 348.52 {\tiny $\pm$ 19.67} & 0.79 {\tiny $\pm$ 0.03}  & 0.98 {\tiny $\pm$ 0.29} & 149.87 {\tiny $\pm$ 4.10} & 18.92 {\tiny $\pm$ 0.94} & 7.83 {\tiny $\pm$ 1.03} \\ \midrule 
PE-RLHF (with MOBIL) & Lateral only & 21.56 {\tiny $\pm$ 8.54} & 368.11 {\tiny $\pm$ 18.45} & 0.81 {\tiny $\pm$ 0.04}  & 0.74 {\tiny $\pm$ 0.19} & 159.34 {\tiny $\pm$ 3.14} & 20.43 {\tiny $\pm$ 0.51} & 9.76 {\tiny $\pm$ 1.17} \\ \midrule
PE-RLHF (with IDM-MOBIL) & Longitudinal \& Lateral  & \textbf{16.61} {\tiny $\pm$ 9.96} & \textbf{391.48 {\tiny $\pm$ 20.47}} & \textbf{0.85} {\tiny $\pm$ 0.04} & \textbf{0.47} {\tiny $\pm$ 0.01} & \textbf{177.00} {\tiny $\pm$ 3.74} & \textbf{21.85} {\tiny $\pm$ 0.02} & \textbf{16.33} {\tiny $\pm$ 4.61} \\ \bottomrule
\end{tabular}%
}
\end{small}
\end{table*}

Tab. \ref{tab2} shows the performance comparison of PE-RLHF with different physics-based model combinations and the standalone IDM-MOBIL model. In Stage I, we observe that PE-RLHF consistently outperforms the standalone IDM-MOBIL model across all configurations. The full PE-RLHF (with IDM-MOBIL) achieves the highest episodic return of 391.48 and a success rate of 0.85, compared to 206.30 and 0.31 for the standalone IDM-MOBIL model, respectively. This substantial improvement demonstrates the effectiveness of integrating RL with physics-based models. Moving to Stage II, we note that all PE-RLHF variants exhibit lower safety violations compared to the standalone IDM-MOBIL model. The full PE-RLHF configuration achieves the lowest safety violation of 0.47, indicating superior safety performance. Additionally, PE-RLHF variants consistently achieve greater travel distances, with the full configuration reaching 177.00m compared to 108.56m for the standalone model. In Stage III,  the full PE-RLHF achieves the highest travel velocity (21.85km/h) and total overtake count (16.33), significantly outperforming the standalone IDM-MOBIL model (19.78 and 0, respectively). 

Interestingly, we observe that incorporating either longitudinal (IDM) or lateral (MOBIL) components of the physics-based model into PE-RLHF yields improvements over the variant without any physics-based model. Yet, the combination of both IDM and MOBIL produces the best results across all metrics, suggesting a synergistic effect when integrating both longitudinal and lateral control models. It is worth noting that while the standalone IDM-MOBIL model provides a baseline level of performance, it struggles with overtaking maneuvers, as evidenced by its zero overtake count. In contrast, all PE-RLHF variants demonstrate the ability to perform overtaking, with the full configuration showing the highest proficiency in this regard. The results demonstrate that the PE-RLHF framework not only leverages the safety guarantees provided by these models but also enhances their performance through learning.

\subsubsection{Impact of Proficiency Level of Human Mentor}

To investigate the impact of human mentor proficiency on the PE-RLHF framework, as shown in Fig. \ref{fig5} (a). we compared two distinct levels of human mentor: professional and amateur. professional mentors are individuals holding Chinese and American passports, with at least 20 hours of cumulative driving experience in the MetaDrive simulator. On the other hand, amateur mentors hold only American passports and have less than 1 hour of driving experience in MetaDrive. 

\begin{figure*}[!t]
  \centering
  \includegraphics[width=0.999\textwidth]{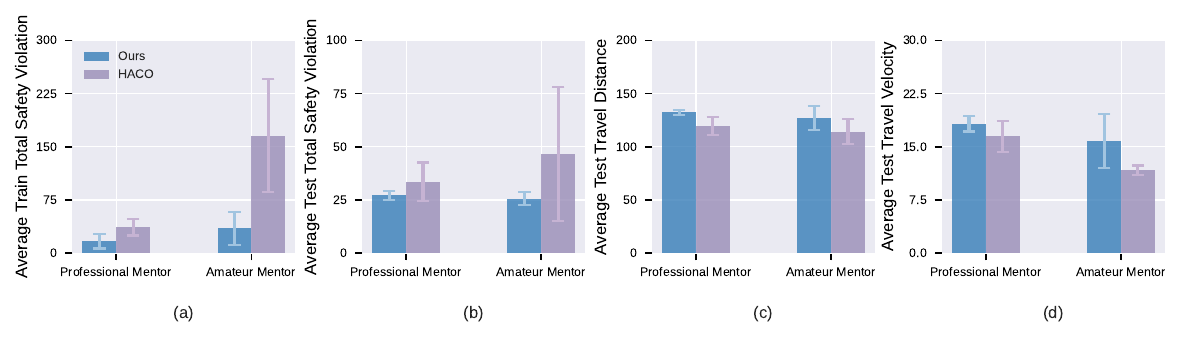}
  \caption{Impact of human mentor proficiency on PE-RLHF performance compared to HACO.}
  \label{fig15}
\end{figure*}

\textit{1) Driving Characteristics.} As shown in Fig. \ref{fig15}, analysis of the driving characteristics reveals significant differences between professional and amateur mentor, while also highlighting the advantages of PE-RLHF across both proficiency levels. The training safety violation for PE-RLHF with professional mentor (16.61) is substantially lower than that of amateur mentor (35.19), indicating a higher level of safety awareness and control. Notably, PE-RLHF significantly improves safety performance for both proficiency levels compared to HACO, with professional mentor achieving a 54.6\% reduction in train safety violation (16.61 vs. 36.59) and amateur mentor showing a 78.7\% reduction (35.19 vs. 165.54). This trend is also reflected in the test safety violation, where PE-RLHF maintains lower average violations for both professional (27.27) and amateur (25.52) mentors compared to HACO (33.48 and 46.48, respectively). The improvement is particularly pronounced for amateur mentor, demonstrating PE-RLHF's ability to mitigate the impact of lower proficiency levels. Furthermore, PE-RLHF enables both mentor groups to achieve greater travel distances and higher velocities. professional mentor using PE-RLHF reached an average distance of 132.44m and velocity of 18.22km/h, outperforming HACO (119.62m and 16.46km/h). Similarly, amateur mentor with PE-RLHF achieve 127.28m and 15.85km/h, surpassing HACO's performance (114.12m and 11.68km/h). These results suggest that PE-RLHF enhances the ability to navigate the environment efficiently while maintaining safety, regardless of mentor proficiency.

\begin{table*}[t]
\centering
\begin{small}
\caption{Performance comparison of PE-RLHF with different human mentor proficiency levels.}
\label{tab3}
\renewcommand{\arraystretch}{2} 
\setlength{\tabcolsep}{1.5pt}
\resizebox{\textwidth}{!}{%
\begin{tabular}{@{}cc|ccc|cc|cc|cc@{}}
\toprule
\multirow{4}{*}{\textbf{Method}} & \multirow{4}{*}{\shortstack{\textbf{Level}}} & \multicolumn{3}{c|}{\multirow{2}{*}{\shortstack{\textbf{Training}}}} & \multicolumn{6}{c}{\textbf{Testing}}  \\ \cline{6-11} & & & & & \multicolumn{2}{c|}{Stage I} & \multicolumn{2}{c|}{Stage II} & \multicolumn{2}{c}{Stage III} \\
\cmidrule(lr){3-5} \cmidrule(lr){6-11} &  & 
\multirow{1}{*}{\shortstack{Total Takeover \\ Data Usage}} \multirow{1}{*}{$\downarrow$} & \multirow{1}{*}{\shortstack{Total Data\\Usage}} \multirow{1}{*}{$\downarrow$} & \multirow{1}{*}{\shortstack{Total Safety\\Violation}} \multirow{1}{*}{$\downarrow$} & \multirow{1}{*}{\shortstack{Episodic\\Return}} \multirow{1}{*}{$\uparrow$}& \multirow{1}{*}{\shortstack{Success Rate \\ (\%)}} \multirow{1}{*}{$\uparrow$} & \multirow{1}{*}{\shortstack{Safety \\ Violation}} \multirow{1}{*}{$\downarrow$} & \multirow{1}{*}{\shortstack{Travel\\Distance}}  \multirow{1}{*}{$\uparrow$}& \multirow{1}{*}{\shortstack{Travel\\Velocity}}  \multirow{1}{*}{$\uparrow$}& \multirow{1}{*}{\shortstack{Total Overtake \\Count}}  \multirow{1}{*}{$\uparrow$} \\
\toprule
\multirow{2}{*}{HACO} & Amateur & 14.40K (0.48) & 30K & 165.54 {\tiny $\pm$79.84}  & 297.40 {\tiny $\pm$8.38} & 0.75 {\tiny $\pm$0.04} & 1.88 {\tiny $\pm$0.41} & 149.77 {\tiny $\pm$12.18} & 13.37 {\tiny $\pm$2.81} & 4.67 {\tiny $\pm$2.08} \\
& Professional & 9.21K (0.31) & 30K &  36.59 {\tiny $\pm$11.64} & 340.73 {\tiny $\pm$10.55} & 0.82 {\tiny $\pm$0.06} & 1.45 {\tiny $\pm$0.98} & 174.28 {\tiny $\pm$14.08}  & 19.22 {\tiny $\pm$3.30} & 7.67 {\tiny $\pm$1.89} \\
\midrule
\multirow{2}{*}{PE-RLHF} & Amateur & 8.81K (0.29) & 30K & 35.19 {\tiny $\pm$23.25} &  376.44 {\tiny $\pm$8.65} & 0.83 {\tiny $\pm$0.05} & 0.77 {\tiny $\pm$0.11} & 176.02 {\tiny $\pm$7.34} & 19.11 {\tiny $\pm$3.89} & 11.67 {\tiny $\pm$1.25} \\
& Professional & 7.86K (0.26)\textsuperscript{*} & 30K & \textbf{16.61} {\tiny $\pm$9.96} & \textbf{ 391.48} {\tiny $\pm$20.47} & \textbf{0.85} {\tiny $\pm$0.04}  & \textbf{0.47} {\tiny $\pm$0.01} & \textbf{177.00} {\tiny $\pm$3.74} & \textbf{21.85} {\tiny $\pm$0.02} & \textbf{16.33} {\tiny $\pm$4.61} \\
\bottomrule
\end{tabular}%
}
\end{small}
\begin{flushleft}
\textsuperscript{*} 
These data represent the average of the maximum values observed at the last checkpoint of each evaluation.
\end{flushleft}
\vspace{-1em}
\vspace{-1em}
\end{table*}

\textit{2) Learning Curves.}
Tab. \ref{tab3} presents a comparison of PE-RLHF performance with different mentor proficiency levels, alongside the baseline HACO method. The results demonstrate that PE-RLHF consistently outperforms HACO across all evaluation phases, regardless of mentor proficiency. In Stage I, PE-RLHF with professional mentor achieves the highest episodic return (391.48) and success rate (0.85), compared to amateur mentor (376.44 and 0.83, respectively). This trend continues in Stage II, where professional mentor enable PE-RLHF to achieve lower safety violations (0.47) and greater travel distances (177.00m) compared to amateur mentor (0.77 and 176.02m, respectively). Stage III metrics further underscore the superiority of professional mentor, with higher travel velocities (21.85km/h) and total overtake counts (16.33) compared to amateur mentor (19.11km/h and 11.67, respectively).

\begin{figure*}[!t]
  \centering
  \includegraphics[width=0.85\textwidth]{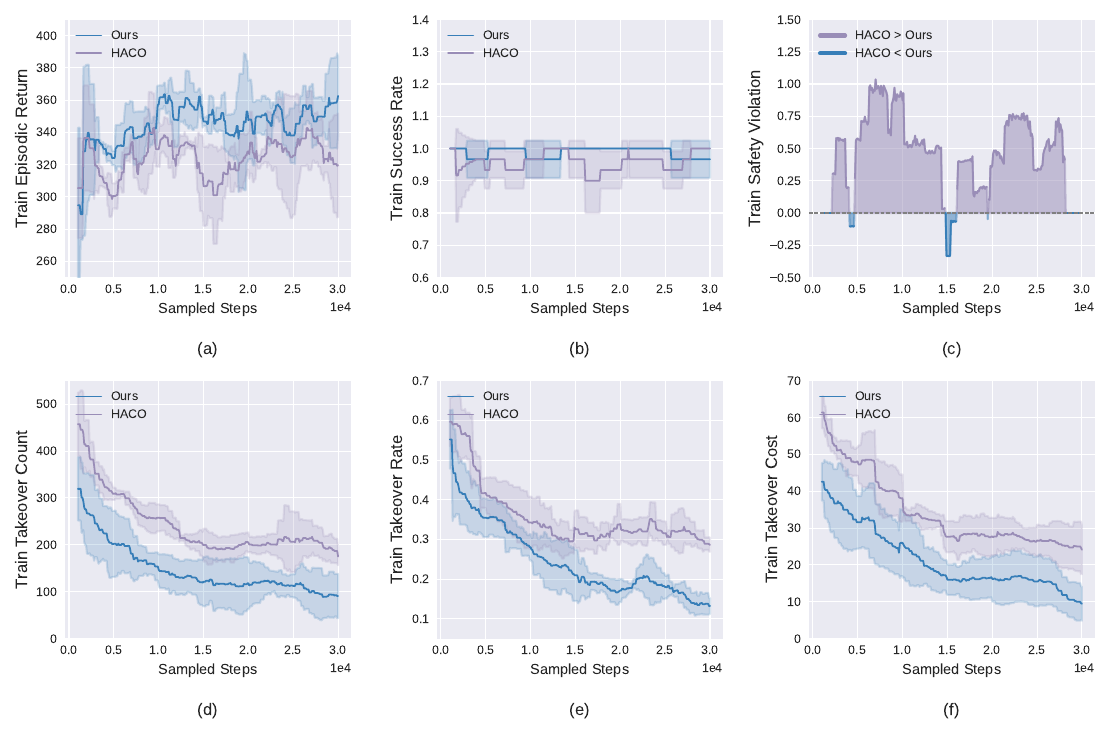}
  \caption{Performance comparison of PE-RLHF with HACO during the training phase by amateur mentor.}
  \label{fig16}
\end{figure*}

\begin{figure*}[!t]
  \centering
  \includegraphics[width=0.85\textwidth]{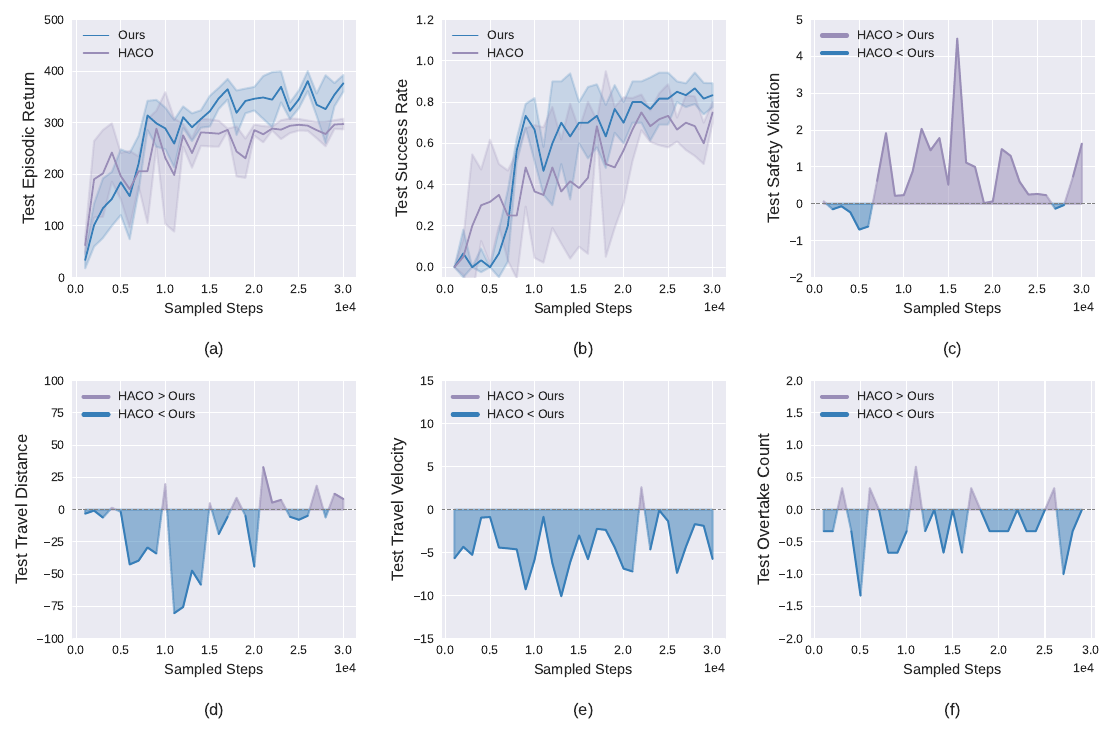}
  \caption{Performance comparison of PE-RLHF with HACO during the testing phase by amateur mentor.}
  \label{fig17}
\end{figure*}

Figs. \ref{fig12}, \ref{fig13}, \ref{fig16}, and \ref{fig17} illustrate the learning curves for professional and amateur mentor, respectively.  Firstly, compared to amateur mentor ( Figs. \ref{fig16}, and \ref{fig17}), professional mentor (Figs. \ref{fig12} and \ref{fig13}) demonstrate faster convergence and more stable performance across all metrics. For example, as shown in Fig. \ref{fig13} (a), PE-RLHF with a professional mentor shows episodic returns reaching a plateau of around 15K steps. In contrast, PE-RLHF with amateur mentor (Fig. \ref{fig17} (a))  shows episodic returns reaching a steady state near 20K steps. However, PE-RLHF outperforms HACO in both scenarios, with HACO exhibiting higher variability and slower convergence, particularly for amateur mentor. The superior convergence characteristics of PE-RLHF, even with amateur mentor, underscore its robustness and ability to effectively leverage both human feedback and physics-based models.

\nopagebreak

\begin{figure*}[!t]
  \centering
  \includegraphics[width=0.75\textwidth]{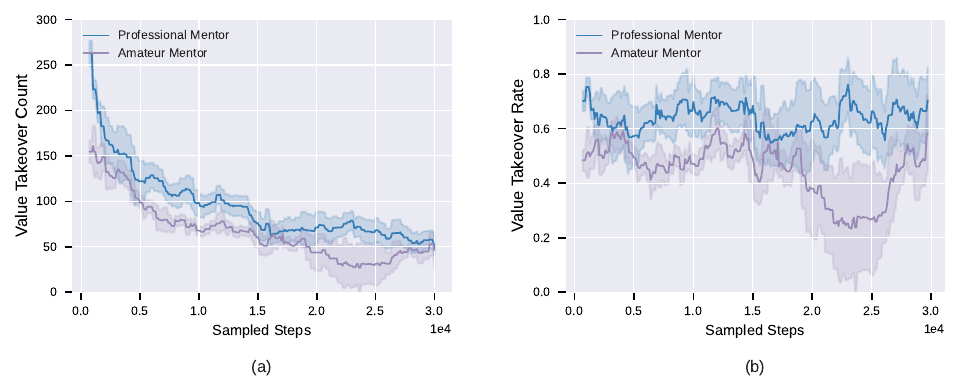}
  \caption{Value takeover analysis for professional and amateur mentor in PE-RLHF.}
  \label{fig18}
\end{figure*}

\textit{3) Value Takeover Analysis.} Fig. \ref{fig18} illustrates the value takeover count and rate for both professional and amateur mentor. The value takeover count represents the number of times the PE-RLHF chooses to use human actions over physics-based model actions during interventions, while the value takeover rate indicates the proportion of human actions used relative to the total number of interventions. Analysis of Fig. \ref{fig18} (a) reveals that PE-RLHF effectively reduces the overall number of value takeovers for both professional and amateur mentor as training progresses. This decrease indicates that PE-RLHF is learning to make better decisions autonomously, reducing the need for human intervention. Notably, the value takeover count for professional mentor consistently remains higher than that of amateur mentor throughout the training process. This observation suggests that PE-RLHF places greater trust in the actions proposed by professional mentor, likely due to their superior action quality and expertise.

Fig. \ref{fig18} (b) provides further insights into the quality of mentor feedback over time. For professional mentor, the value takeover rate remains relatively stable, hovering around 0.6-0.7 throughout the training process. In contrast, amateur mentor exhibit a declining value takeover rate, starting at a similar level to professional mentor but dropping significantly to nearly zero around the 25K step mark. This decline suggests that the quality of amateur mentor' feedback deteriorates over time, possibly due to factors such as fatigue or inconsistency. Despite the apparent decline in amateur mentor' feedback quality, Tab. \ref{tab3} demonstrates that PE-RLHF still achieves superior performance compared to HACO for both mentor types. This resilience can be attributed to PE-RLHF's integration of physics-based models as a safety guarantee. The framework shows particular strength in maintaining high performance even when faced with declining feedback quality from amateur mentor.

\subsubsection{Impact of Warmup Strategies}

\begin{figure*}[h]
  \centering
  \includegraphics[width=0.85\textwidth]{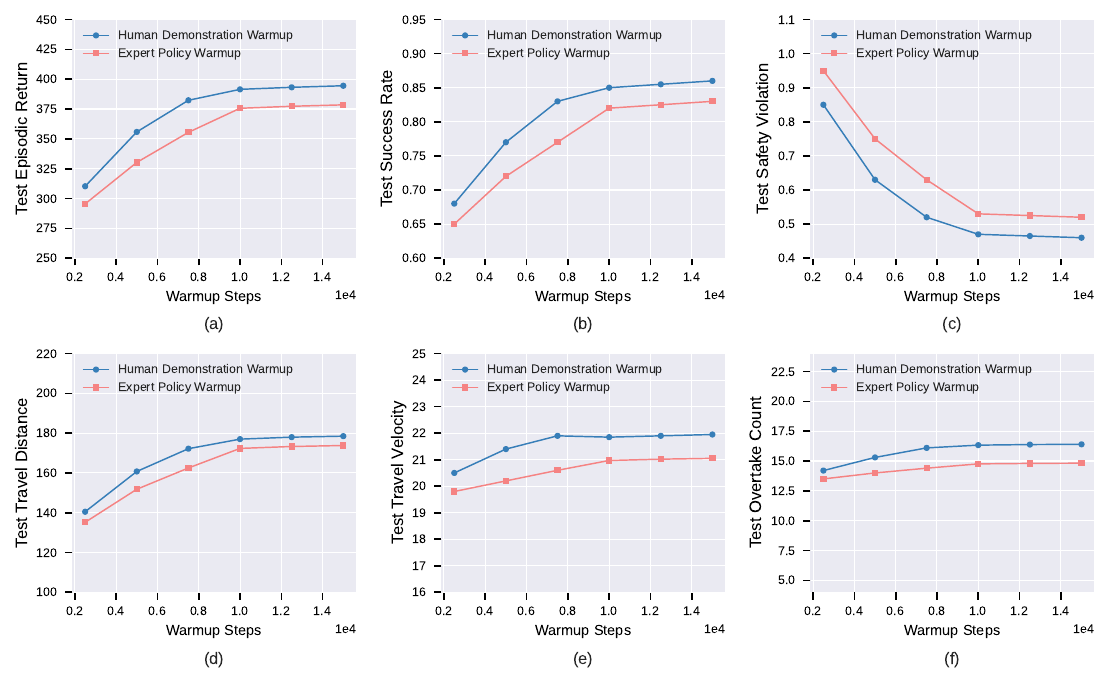}
  \caption{Performance comparison of different warmup strategies for PE-RLHF.}
  \label{fig19}
\end{figure*}

To investigate the impact of different warmup strategies, we compared human demonstration warmup and expert policy warmup, the results shown in Fig. \ref{fig19}. The comparison was conducted over a range of warmup steps from 2.5K to 15K. In Stage I, human demonstration warmup consistently outperforms expert policy warmup in terms of episodic return and success rate. The gap is particularly pronounced in the early stages and remains significant even as the number of warmup steps increases. At 15K steps, human demonstration warmup achieves an episodic return of approximately 394 and a success rate of about 0.86, compared to 378 and 0.83 for expert policy warmup, respectively. Stage II metrics also favor human demonstration warmup. It maintains lower safety violation rates and achieves longer travel distances across all warmup steps. At 15K steps, human demonstration warmup reaches a safety violation rate of about 0.46 and a travel distance of approximately 180m, while expert policy warmup achieves 0.52 and 173.8m, respectively. In Stage III, human demonstration warmup again demonstrates superiority. While travel velocity profiles are similar for both methods, human demonstration warmup maintains a slight edge. More significantly, it leads to higher overtake counts across all warmup steps, indicating more dynamic and efficient driving behavior. The performance gap between the two strategies tends to narrow as the number of warmup steps increases, particularly after 10K steps. Considering the diminishing returns in performance improvement beyond 10K steps, this study determines that selecting 10K warmup steps is a reasonable choice for the PE-RLHF framework. This decision optimizes the trade-off between training efficiency and performance, ensuring that the system achieves high performance without unnecessary computational or human resource expenditure.

\subsubsection{Value Takeover Threshold}

\begin{figure*}[h]
  \centering
  \includegraphics[width=0.85\textwidth]{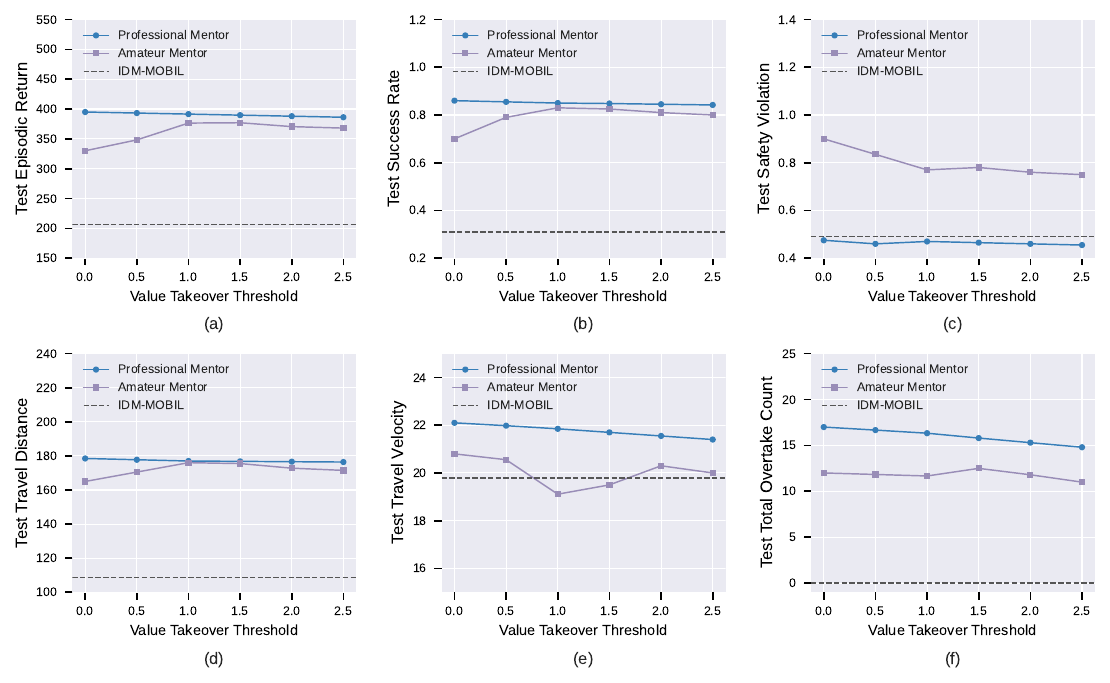}
  \caption{Sensitivity analysis of value takeover threshold in PE-RLHF for different mentor proficiency levels.}
  \label{fig20}
\end{figure*}

We conducted a sensitivity analysis by varying $\varepsilon_{\text{select}} $ in Eq. \ref{eq17} and observing its effects on the framework's performance across different proficiency levels of human mentors. Fig. \ref{fig20} illustrates the performance metrics for professional mentor, amateur mentor, and the standalone IDM-MOBIL model as $\varepsilon_{\text{select}} $ increases. In Stage I, we observe that both professional and amateur mentor consistently outperform the IDM-MOBIL model in terms of episodic return and success rate across all $\varepsilon_{\text{select}} $ values. This suggests that the PE-RLHF framework effectively leverages human feedback, regardless of the mentor's proficiency level. The performance gap between professional and amateur mentor is more pronounced at lower $\varepsilon_{\text{select}} $ values but narrows as $\varepsilon_{\text{select}} $ increases. In Stage II, we observe that PE-RLHF achieves lower safety violations compared to the IDM-MOBIL model for both professional and amateur mentor types. In Stage III, the travel velocity and total overtake count demonstrate PE-RLHF's ability to learn more advanced driving behaviors. Notably, professional mentor facilitate a higher overtake count, suggesting that the framework can capitalize on high-quality human feedback for complex driving scenarios, especially at lower $\varepsilon_{\text{select}} $ values. As $\varepsilon_{\text{select}} $ increases, we observe that the performance gap between professional and amateur mentor generally narrows across all metrics. This suggests that at higher $\varepsilon_{\text{select}} $ values, the framework relies more heavily on the physics-based model, which helps to compensate for variations in human feedback quality. Conversely, at lower $\varepsilon_{\text{select}} $ values, the PE-RLHF makes greater use of human feedback, allowing the superior performance of a professional mentor to be more evident.

\subsection{Ablation Study}

To evaluate the contribution of each component in the proposed PE-RLHF framework, we conducted an ablation study. Tab. \ref{tab4} presents the results of this ablation study, comparing the full PE-RLHF model against variants with specific components removed or modified. 

\subsubsection{Frequency of human takeover}
Reducing the frequency of human takeover resulted in a significant performance decrease across all metrics. The episodic return dropped to 248.36, and the success rate declined to 0.58. This variant also showed higher safety violations (2.41) and lower travel distances (135.24m). The decrease in performance highlights the crucial role of timely human intervention in the learning process. The reduced performance suggests that frequent human feedback provides valuable guidance and correction. The higher safety violation indicates that less frequent human intervention may lead to riskier decisions by the AV. Furthermore, the lower travel distance implies that the AV becomes more conservative or less efficient in its navigation without regular human input.

\subsubsection{Cosine similarity takeover cost}
Removing the cosine similarity takeover cost led to degraded performance, particularly in safety-related metrics. The episodic return fell to 172.15, with a success rate of only 0.46. Notably, the number of safety violations increased to 2.93, indicating a significant reduction in driving safety. The travel velocity and overtake count also decreased to 14.12km/h and 4.29 respectively. These results suggest that without a cosine similarity component, the framework may struggle to integrate human feedback effectively. The decreased overtake count suggests that the framework becomes more conservative when in complex maneuvers without this smoothing mechanism, potentially due to increased uncertainty in action selection.

\begin{table}[t]
\centering
\begin{small}
\caption{Ablation study results for PE-RLHF.}
\label{tab4}
\renewcommand{\arraystretch}{1.45} 
\setlength{\tabcolsep}{0.5pt} 
\begin{tabular}{@{}>{\centering\arraybackslash}m{5cm}|cc|cc|cc@{}}
\toprule
\multirow{3}{*}{\textbf{Experiment}} & \multicolumn{6}{c}{\shortstack{\textbf{Testing}}}  \\\cline{2-7}  & \multicolumn{2}{c|}{Stage I} & \multicolumn{2}{c|}{Stage II} & \multicolumn{2}{c}{Stage III} \\
\cmidrule(lr){2-3} \cmidrule(lr){4-5} \cmidrule(lr){6-7}
&  \multirow{1}{*}{\shortstack{Episodic\\Return}}  \multirow{1}{*}{$\uparrow$}  & \multirow{1}{*}{\shortstack{Success\\Rate}}  \multirow{1}{*}{$\uparrow$}  & \multirow{1}{*}{\shortstack{Safety \\ Violation}}  \multirow{1}{*}{$\downarrow$} & \multirow{1}{*}{\shortstack{Travel\\Distance}}  \multirow{1}{*}{$\uparrow$}  & \multirow{1}{*}{\shortstack{Travel\\Velocity}}  \multirow{1}{*}{$\uparrow$}  & \multirow{1}{*}{\shortstack{Total Overtake\\Count}}  \multirow{1}{*}{$\uparrow$}  \\ \midrule
\raggedright Human takeover less frequently & 248.36 & 0.58 & 2.41 & 135.24 & 13.67 & 5.23 \\ 
\midrule
\raggedright W/o cosine similarity takeover cost & 172.15 & 0.46 & 2.93 & 127.89 & 14.12 & 4.29 \\ 
\midrule
\raggedright W/o intervention minimization & 156.78 & 0.32 & 3.67 & 113.45 & 12.32 & 3.57 \\ 
\midrule
\raggedright W/o Ensemble technique & 340.25 & 0.80 & 0.89 & 165.32 & 19.76 & 12.45 \\ 
\midrule
\raggedright PE-RLHF (Full) & \textbf{391.48} & \textbf{0.85} & \textbf{0.47} & \textbf{177.00} & \textbf{21.85} & \textbf{16.33}  \\ \bottomrule
\end{tabular}
\end{small}
\end{table}

\subsubsection{Intervention minimization}
The variant without intervention minimization exhibited the poorest performance across nearly all metrics. It achieved the lowest episodic return (156.78) and success rate (0.32) among all variants. The safety violation was highest (3.67), and the travel distance was shortest (113.45m). Additionally, this variant showed the lowest travel velocity (12.32km/h) and overtake count (3.57). These results strongly suggest that intervention minimization is critical for balancing the trade-off between leveraging human mentors and developing autonomous capabilities. Without this component, the framework likely becomes overly dependent on human feedback and fails to develop robust autonomous decision-making skills.

\subsubsection{Ensemble Technique}
While the variant without the ensemble technique still performed well, achieving an episodic return of 340.25 and a success rate of 0.78, the full model including the ensemble technique showed marked improvements across all metrics. Furthermore, the improvements in safety violation and travel distance highlight the ensemble technique's contribution to both safety and efficiency. These results confirm that the ensemble technique successfully achieves its primary goal of enhancing the robustness of the value estimator, particularly in previously unseen states, thereby improving the overall decision-making process of the PE-RLHF framework.

\section{Conclusions and Future Work}
\label{Conclusions and Future Work}

In this paper, we proposed \textbf{Physics-enhanced Reinforcement Learning with Human Feedback (PE-RLHF)}, a novel framework for trustworthy decision-making in autonomous vehicles. To the best of the author's knowledge, PE-RLHF is the first reinforcement learning (RL)-based framework that simultaneously incorporates human feedback (e.g., human intervention and demonstration) and physics knowledge (e.g., traffic flow model) for driving policy learning. The key innovation of PE-RLHF is that it estimates the value of both human actions and physics-based policies during the RL iteration process, ensuring that the final action executed by the agent has the higher policy value between the two. 

Our comprehensive evaluation, comparing PE-RLHF with traditional physics-based methods, various RL approaches, and existing RLHF methods, yielded significant insights: (a) PE-RLHF demonstrated trustworthy safety improvement, achieving a safety violation of 0.47, which significantly outperforms the standalone IDM-MOBIL model and other advance RLHF methods. (b) PE-RLHF exhibited high sampling efficiency, reducing the required training data by 74\% and training time from over 30 hours to just 1 hour compared to traditional RL methods. (c) PE-RLHF showed robustness to varying human feedback quality, maintaining performance at least as good as existing interpretable physics-based models, even when human feedback quality deteriorated.

The broader impact of PE-RLHF extends beyond autonomous driving, potentially influencing fields such as robotics, manufacturing, and smart city management. By synergistically integrating human expertise, physical knowledge, and artificial intelligence, PE-RLHF opens new avenues for developing more reliable AI systems in safety-critical applications. While PE-RLHF demonstrates significant improvements over traditional methods, it is important to acknowledge its reliance on certain assumptions and limitations. Future research directions to enhance PE-RLHF include learning from diverse human experts to account for varying driving styles and cultural contexts, integrating large language models (LLMs) for improved contextual understanding and decision-making, and conducting extensive real-world tests to validate the framework's performance in diverse and unpredictable environments.

\section*{Acknowledgment}
This work was supported by the University of Wisconsin-Madison’s Center for Connected and Automated Transportation (CCAT), a part of the larger CCAT consortium, a USDOT Region 5 University Transportation Center funded by the U.S. Department of Transportation, Award \#69A3552348305. The contents of this paper reflect the views of the authors, who are responsible for the facts and the accuracy of the data presented herein, and do not necessarily reflect the official views or policies of the sponsoring organization.
	
\appendix

\section{Proof of Theorem \ref{theorem1}}
\label{Appendix A}

\noindent \textbf{Lemma A.1} Lemma 4.1 in \citep{xu2019value}
\begin{equation}
\left\|d_{\pi}-d_{\pi'}\right\|_1 \leqslant \frac{\gamma}{1-\gamma} \mathbb{E}_{s \sim d_{\pi}}\left\|\pi(\cdot \mid s)-\pi'(\cdot \mid s)\right\|_1.
\end{equation}

\noindent Lem. \ref{lemma1} can be derived by substituting $\pi$ and $\pi'$ in Lemma 1 with $\pi_{\text{human}}$ and $\pi_{\text{AV}}$.
\vspace{0.2cm}

\noindent \textbf{Theorem A.2} (Restatement of Thm. \ref{theorem1}). \textit{For any mixed behavior policy $\pi_{\text{mix}}$ deduced by a human policy $\pi_{\text{human}}$, an AV policy $\pi_{\text{AV}}$, and a switch function $\mathcal{T}(s)$, the state distribution discrepancy between $\pi_{\text{mix}}$ and $\pi_{\text{AV}}$ is bounded by policy discrepancy and intervention rate:}
\begin{equation}
\left\|d_{\pi_{\text{mix}}}- d_{\pi_{\text{AV}}}\right\|_{1}\leqslant
\frac{\beta\gamma}{1-\gamma} \mathbb{E}_{s \sim d_{\pi_{\text{mix}}}} \left\|\pi_{\text{human}}(\cdot \mid s)- \pi_{\text{AV}}(\cdot \mid s)\right\|_{1},
\end{equation}
where $\beta=\frac{\mathbb{E}_{s \sim d_{\pi_{\text{mix}}}}\left\|\mathcal{T}(s)\left[\pi_{\text{human}}(\cdot \mid s)-\pi_{\text{AV}}(\cdot \mid s)\right]\right\||_{1}}{\mathbb{E}_{s \sim d_{\pi_{\text{mix}}}}\left\|\pi_{\text{human}}(\cdot \mid s)-\pi_{\text{AV}}(\cdot \mid s)\right\|_{1}}$ is the weighted expected intervention rate.
\vspace{0.2cm}

\noindent \textit{Proof.}
\begin{equation}
\begin{aligned}
\left\|d_{\pi_{\text{mix}}}-d_{\pi_{\text{AV}}}\right\|_1 &\leqslant \frac{\gamma}{1-\gamma} \mathbb{E}_{s \sim d_{\pi_{\text{mix}}}}\left\|\pi_{\text{mix}}(\cdot \mid s)-\pi_{\text{AV}}(\cdot \mid s)\right\|_1\\
&=\frac{\gamma}{1-\gamma} \mathbb{E}_{s \sim d_{\pi_{\text{mix}}}}\left\|\mathcal{T}(s)\pi_{\text{human}}(\cdot \mid s)+(1-\mathcal{T}(s))\pi_{\text{AV}}(\cdot \mid s)-\pi_{\text{AV}}(\cdot \mid s)\right\|_1\\
&=\frac{\gamma}{1-\gamma} \mathbb{E}_{s \sim d_{\pi_{\text{mix}}}}\left\|\mathcal{T}(s)\left[\pi_{\text{human}}(\cdot \mid s)-\pi_{\text{AV}}(\cdot \mid s)\right]\right\|_1\\
&=\frac{\beta\gamma}{1-\gamma} \mathbb{E}_{s \sim d_{\pi_{\text{mix}}}} \left\|\pi_{\text{human}}(\cdot \mid s)- \pi_{\text{AV}}(\cdot \mid s)\right\|_1.
    \end{aligned}
\end{equation}
\vspace{0.2cm}

\noindent Based on Thm. \ref{theorem1}, we further prove that under the setting of PE-HAI, the performance gap between the $\pi_{\text{AV}}$ and the optimal AV policy $\pi_{\text{AV}}^*$ can be upper bounded by the combination of two terms: (1) the gap between the $\pi_{\text{human}}$ and the $\pi_{\text{AV}}^*$, and (2) the discrepancy between the $\pi_{\text{human}}$ and the $\pi_{\text{AV}}$. Consequently, training the $\pi_{\text{AV}}$ using trajectories collected by the $\pi_{\text{mix}}$ essentially optimizes an upper bound of the AV's suboptimality, which provides a principled approach to reduce the performance gap between the learned AV policy and the optimal AV policy in the presence of human interventions. The following lemma helps prove this result.
\vspace{0.2cm}

\noindent \textbf{Lemma A.3} 
\begin{equation}
\left|J\left(\pi\right)-J\left(\pi^\prime\right)\right| \leqslant \frac{R_{\max}}{(1-\gamma)^2}\mathbb{E}_{s \sim d_{\pi}} \left\|\pi(\cdot \mid s)- \pi^\prime(\cdot \mid s)\right\|_1
\end{equation}
\vspace{0.2cm}

\noindent \textit{Proof.} The proof follows directly from the combination of Lemma 4.2 and Lemma 4.3 in \citep{xu2019value}.
\vspace{0.2cm}

\noindent \textbf{Theorem A.4} \textit{For any mixed behavior policy $\pi_{\text{mix}}$ consisting of a human policy $\pi_{\text{human}}$, an AV policy $\pi_{\text{AV}}$, and a switch function $\mathcal{T}(s)$, the suboptimality of the AV policy is bounded by}
\vspace{0.2cm}

\begin{equation}
\begin{aligned}
    \left|J\left(\pi_{\text{AV}}^*\right)-J\left(\pi_{\text{AV}}\right)\right| \leqslant& \frac{\beta R_{\max}}{(1-\gamma)^{2}} \mathbb{E}_{s\sim \pi_{\text{mix}}}\left\|\pi_{\text{human}}(\cdot \mid s)- \pi_{\text{AV}}(\cdot \mid s)\right\|_1+\left|J\left(\pi_{\text{AV}}^*\right)-J\left(\pi_{\text{mix}}\right)\right| ,
\end{aligned}
\end{equation}
\vspace{0.2cm}

\noindent \textit{Proof.}
\begin{equation}
\begin{aligned}
        \left|J\left(\pi_{\text{mix}}\right)-J\left(\pi_{\text{AV}}\right)\right| &\leqslant \frac{R_{\max}}{(1-\gamma)^2}\mathbb{E}_{s \sim d_{\pi_{\text{mix}}}} \left\|\pi_{\text{mix}}(\cdot \mid s)- \pi_{\text{AV}}(\cdot \mid s)\right\|_1\\
        &=\frac{R_{\max}}{(1-\gamma)^2}\mathbb{E}_{s \sim d_{\pi_{\text{mix}}}}\left\|\mathcal{T}(s)\pi_{\text{human}}(\cdot \mid s)+(1-\mathcal{T}(s))\pi_{\text{AV}}(\cdot \mid s)-\pi_{\text{AV}}(\cdot \mid s)\right\|_1\\
        &=\frac{R_{\max}}{(1-\gamma)^2}\mathbb{E}_{s \sim d_{\pi_{\text{mix}}}}\left\|\mathcal{T}(s)\left[\pi_{\text{human}}(\cdot \mid s)-\pi_{\text{AV}}(\cdot \mid s)\right]\right\|_1\\
        &=\frac{\beta R_{\max}}{(1-\gamma)^{2}} \mathbb{E}_{s\sim \pi_{\text{mix}}}\left\|\pi_{\text{human}}(\cdot \mid s)- \pi_{\text{AV}}(\cdot \mid s)\right\|_1.\\
        \left|J\left(\pi_{\text{AV}}^*\right)-J\left(\pi_{\text{AV}}\right)\right|&\leqslant\left|J\left(\pi_{\text{mix}}\right)-J\left(\pi_{\text{AV}}\right)\right|+\left|J\left(\pi_{\text{AV}}^*\right)-J\left(\pi_{\text{mix}}\right)\right|\\
        &\leqslant\frac{\beta R_{\max}}{(1-\gamma)^{2}} \mathbb{E}_{s\sim \pi_{\text{mix}}}\left\|\pi_{\text{human}}(\cdot \mid s)- \pi_{\text{AV}}(\cdot \mid s)\right\|_1+\left|J\left(\pi_{\text{AV}}^*\right)-J\left(\pi_{\text{mix}}\right)\right|.
    \end{aligned}
\end{equation}

\section{Proof of  Theorem \ref{theorem2}}
\label{Appendix B}

\noindent \textbf{Theorem B.1} (Restatement of Thm. \ref{theorem2}). \textit{With the action-based switch function $\mathcal{T}_\text{action}(s)$, the return of the mixed behavior policy $J(\pi_{\text{mix}})$ is lower and upper bounded by}
\begin{equation}
\begin{aligned}
J(\pi_{\text{human}})+&\frac{\sqrt{2}(1-\beta) R_{\max }}{(1-\gamma)^{2}} \sqrt{H-\varepsilon}\geqslant J(\pi_{\text{mix}})\geqslant J(\pi_{\text{human}})-\frac{\sqrt{2}(1-\beta) R_{\max }}{(1-\gamma)^{2}} \sqrt{H-\varepsilon}
\end{aligned}
\end{equation}
where $R_{\max}=\max\limits_{s,a}r(s,a)$ is the maximal possible reward, $H=\mathbb{E}_{s\sim d_{\pi_{\text{mix}}}} \mathcal{H}(\pi^t(\cdot|s))$ is the average entropy of the human policy during shared control.
\vspace{0.2cm}

\noindent \textit{Proof.}
\begin{equation}
\begin{aligned}
     \left|J\left(\pi_{\text{mix}}\right)-J\left(\pi_{\text{human}}\right)\right| &\leqslant\frac{R_{\max}}{(1-\gamma)^2}\mathbb{E}_{s \sim d_{\pi_{\text{mix}}}} \left\|\pi_{\text{mix}}(\cdot \mid s)- \pi_{\text{human}}(\cdot \mid s)\right\|_1\\
     &=\frac{R_{\max}}{(1-\gamma)^2}\mathbb{E}_{s \sim d_{\pi_{\text{mix}}}} \left\|\mathcal{T}(s)\pi_{\text{human}}(\cdot \mid s)+(1-\mathcal{T}(s))\pi_{\text{AV}}(\cdot \mid s)-\pi_{\text{human}}(\cdot \mid s)\right\|_1\\
     &=\frac{(1-\beta)R_{\max}}{(1-\gamma)^2}\mathbb{E}_{s \sim d_{\pi_{\text{mix}}}} \left\|\pi_{\text{AV}}(\cdot \mid s)- \pi_{\text{human}}(\cdot \mid s)\right\|_1\\
     &\leqslant\frac{\sqrt{2}(1-\beta)R_{\max}}{(1-\gamma)^2}\mathbb{E}_{s \sim d_{\pi_{\text{mix}}}}\sqrt{\mathrm{D}_{\mathrm{KL}}(\pi_{\text{human}}(\cdot|s)\|\pi_{\text{AV}}(\cdot|s))}\\
     &=\frac{\sqrt{2}(1-\beta)R_{\max}}{(1-\gamma)^2}\mathbb{E}_{s \sim d_{\pi_{\text{mix}}}}\sqrt{\mathbb{E}_{a\sim\pi_{\text{human}}(\cdot|s)}\left[\log\pi_{\text{human}}(a|s)-\log\pi_{\text{AV}}(a|s)\right]}\\
     &=\frac{\sqrt{2}(1-\beta)R_{\max}}{(1-\gamma)^2}\mathbb{E}_{s \sim d_{\pi_{\text{mix}}}}\sqrt{\mathcal{H}(\pi_{\text{human}}(\cdot|s)-\varepsilon}\\
     &\leqslant\frac{\sqrt{2}(1-\beta)R_{\max}}{(1-\gamma)^2}\sqrt{H-\varepsilon}.
\end{aligned}
\end{equation}
\vspace{0.2cm}

\noindent Therefore, we obtain
\begin{equation}
\begin{aligned}
    \frac{\sqrt{2}(1-\beta)R_{\max}}{(1-\gamma)^2}\sqrt{H-\varepsilon}\geqslant J\left(\pi_{\text{mix}}\right)&-J\left(\pi_{\text{human}}\right)\geqslant -\frac{\sqrt{2}(1-\beta)R_{\max}}{(1-\gamma)^2}\sqrt{H-\varepsilon}\\
    J(\pi_{\text{human}})+\frac{\sqrt{2}(1-\beta) R_{\max }}{(1-\gamma)^{2}} \sqrt{H-\varepsilon}\geqslant &J(\pi_{\text{mix}})\geqslant J(\pi_{\text{human}})-\frac{\sqrt{2}(1-\beta) R_{\max }}{(1-\gamma)^{2}} \sqrt{H-\varepsilon},
\end{aligned}
\end{equation}
\noindent which concludes the proof.
\vspace{0.2cm}

\section{Proof of Theorem \ref{theorem3}}
\label{Appendix C}
\noindent \textbf{Theorem C.1} (Restatement of Thm. \ref{theorem3}). \textit{The expected cumulative reward obtained by learning from $\pi_{\text{hybrid}}$ is equal to the maximum of the expected cumulative rewards obtained by the $\pi_{\text{human}}$ and the $\pi_{\text{phy}}$. It is also guaranteed to be greater than or equal to the expected cumulative reward obtained by the $\pi_{\text{phy}}$. }
\begin{equation}
\label{eqc1}
\begin{aligned}
\mathbb{E}_{\tau \sim \pi_{\text{hybrid}}}\left[\sum_{t=0}^{H} \gamma^t r(s_t, a_t)\right] & = \max\left(\mathbb{E}_{\tau \sim \pi_{\text{human}}}\left[\sum_{t=0}^{H} \gamma^t r(s_t, a_t)\right], \mathbb{E}_{\tau \sim \pi_{\text{phy}}}\left[\sum_{t=0}^{H} \gamma^t r(s_t, a_t)\right]\right) \\
& \geq \mathbb{E}_{\tau \sim \pi_{\text{phy}}}\left[\sum_{t=0}^{H} \gamma^t r(s_t, a_t)\right]
\end{aligned}
\end{equation}

\noindent \textit{Proof.}
\noindent We will prove this theorem by induction on the decision horizon. Let us begin by considering the simplest case where the horizon $H = 0$, meaning only one decision is made. In this scenario, the hybrid policy selects the action that yields the highest immediate reward:
\begin{equation}
\pi_{\text{hybrid}}(s) = \begin{cases}
\pi_{\text{human}}(s) & \text{if } r(s, \pi_{\text{human}}(s)) \geq r(s, \pi_{\text{phy}}(s)) \\
\pi_{\text{phy}}(s) & \text{otherwise}
\end{cases}
\end{equation}

\noindent Therefore, for $H = 0$, we have:
\begin{equation}
\mathbb{E}_{\tau \sim \pi_{\text{hybrid}}}[r(s, a)] = \max(\mathbb{E}_{\tau \sim \pi_{\text{human}}}[r(s, a)], \mathbb{E}_{\tau \sim \pi_{\text{phy}}}[r(s, a)])
\end{equation}

\noindent Now, let us assume that the theorem holds for some horizon $k \geq 0$. That is:
\begin{equation}
\mathbb{E}_{\tau \sim \pi_{\text{hybrid}}}\left[\sum_{t=0}^{k} \gamma^t r(s_t, a_t)\right] = \max\left(\mathbb{E}_{\tau \sim \pi_{\text{human}}}\left[\sum_{t=0}^{k} \gamma^t r(s_t, a_t)\right], \mathbb{E}_{\tau \sim \pi_{\text{phy}}}\left[\sum_{t=0}^{k} \gamma^t r(s_t, a_t)\right]\right)
\end{equation}

\noindent With this assumption in place, we need to demonstrate that the theorem holds for horizon $k+1$. To do this, let us define the value function for a policy $\pi$ as:
\begin{equation}
V_{\pi}(s) = \mathbb{E}_{\tau \sim \pi}\left[\sum_{t=0}^{k+1} \gamma^t r(s_t, a_t) \mid s_0 = s\right]
\end{equation}

\noindent Using the Bellman equation, we can express this as:
\begin{equation}
V_{\pi}(s) = r(s, \pi(s)) + \gamma \mathbb{E}_{s' \sim P(s'|s,\pi(s))}[V_{\pi}(s')]
\end{equation}

\noindent For our hybrid policy, this equation becomes:
\begin{equation}
V_{\pi_{\text{hybrid}}}(s) = \max\left(r(s, \pi_{\text{human}}(s)) + \gamma \mathbb{E}_{s'}[V_{\pi_{\text{hybrid}}}(s')], r(s, \pi_{\text{phy}}(s)) + \gamma \mathbb{E}_{s'}[V_{\pi_{\text{hybrid}}}(s')]\right)
\end{equation}

\noindent By the inductive hypothesis, we know that $V_{\pi_{\text{hybrid}}}(s') = \max(V_{\pi_{\text{human}}}(s'), V_{\pi_{\text{phy}}}(s'))$ for all $s'$. Substituting this into the above equation:
\begin{align}
V_{\pi_{\text{hybrid}}}(s) &= \max\left(r(s, \pi_{\text{human}}(s)) + \gamma \mathbb{E}_{s'}[\max(V_{\pi_{\text{human}}}(s'), V_{\pi_{\text{phy}}}(s'))], \right. \\
&\qquad \left. r(s, \pi_{\text{phy}}(s)) + \gamma \mathbb{E}_{s'}[\max(V_{\pi_{\text{human}}}(s'), V_{\pi_{\text{phy}}}(s'))]\right) \\
&\geq \max\left(r(s, \pi_{\text{human}}(s)) + \gamma \mathbb{E}_{s'}[V_{\pi_{\text{human}}}(s')], r(s, \pi_{\text{phy}}(s)) + \gamma \mathbb{E}_{s'}[V_{\pi_{\text{phy}}}(s')]\right) \\
&= \max(V_{\pi_{\text{human}}}(s), V_{\pi_{\text{phy}}}(s)) \geq V_{\pi_{\text{phy}}}(s)
\end{align}

\noindent This result demonstrates that the theorem holds for horizon $k+1$. By the principle of mathematical induction, we can conclude that the theorem holds for all finite horizons $H \geq 0$.

\noindent Thus, the hybrid policy $\pi_\text{hybrid}$ outperforms the existing physics-based policy $\pi_\text{phy}$, namely:
\begin{equation}
\label{eqc15}
\begin{aligned}
\mathbb{E}_{\tau \sim \pi_{\text{hybrid}}}\left[\sum_{t=0}^{H} \gamma^t r(s_t, a_t)\right] \geq \mathbb{E}_{\tau \sim \pi_{\text{phy}}}\left[\sum_{t=0}^{H} \gamma^t r(s_t, a_t)\right]
\end{aligned}
\end{equation}
\vspace{0.2cm}

\noindent Theorem \ref{theorem3} is proved.


\section{Workflow of PE-HAI Collaborative Paradigm}
\label{Appendix:Workflow-PE-HAI}
\begin{algorithm}[H]
\caption{PE-HAI Collaborative Paradigm}
\begin{algorithmic}[1]
\Require{Initial AV policy $\pi_{\text{AV}}$, human policy $\pi_{\text{human}}$, physics-based policy $\pi_{\text{phy}}$, switch function $\mathcal{T}(s)$, selection function $\mathcal{T}_{\text{select}}(s)$, ensemble of Q-networks $\mathbf{Q}^\phi$, threshold $\varepsilon_{\text{select}}$}
\For{each episode}
\For{each step $t$}
\State{Observe state $s_t$}
\If{$\mathcal{T}(s_t) = 1$} // Human takeover
\State{$a_t^{\text{human}} \sim \pi_{\text{human}}(\cdot \mid s_t)$}
\State{$a_t^{\text{phy}} \leftarrow \pi_{\text{phy}}(s_t)$}
\If{$\operatorname{Mean}\left[\mathbb{E}_{a \sim \pi_{\text{human}}(\cdot \mid s_t)} \mathbf{Q}^\phi(s_t, a)-\mathbb{E}_{a \sim \pi_{\text{phy}}(\cdot \mid s_t)} \mathbf{Q}^\phi(s_t, a)\right] \geq \varepsilon_{\text{select}}$}
\State{$a_t \leftarrow a_t^{\text{human}}$} // Execute human policy action
\Else
\State{$a_t \leftarrow a_t^{\text{phy}}$} // Execute physics-based policy action
\EndIf
\Else 
\State{$a_t \sim \pi_{\text{AV}}(\cdot \mid s_t)$} // Execute AV action
\EndIf
\State{Execute action $a_t$ and observe next state $s_{t+1}$}
\State{Store transition $(s_t, a_t, s_{t+1})$ in replay buffer $\mathcal{D}$}
\State{Update $\pi_{\text{AV}}$ using transitions from $\mathcal{D}$}
\EndFor
\EndFor
\end{algorithmic}
\end{algorithm}

\section{Convergence Analysis of Proxy Q Function}
\label{Appendix D}
\noindent \textbf{Lemma D.1 (Contraction Property of the Bellman Operator)} \textit{
The Bellman operator $\mathcal{T}^{\pi}$ for the proxy $Q$ function under policy $\pi$, defined as}
\begin{equation}
\mathcal{T}^{\pi} \hat{Q}(s,a) = \gamma \mathbb{E}{s' \sim P(\cdot|s,a)} \left[ \max{a'} \hat{Q}(s',a') \right],
\end{equation}
\vspace{0.2cm}
\noindent is a contraction mapping concerning the maximum norm \citep{sutton2018reinforcement}, i.e.,
\begin{equation}
|\mathcal{T}^{\pi} \hat{Q}_1 - \mathcal{T}^{\pi} \hat{Q}2|{\infty} \leq \gamma |\hat{Q}_1 - \hat{Q}2|{\infty}.
\end{equation}
\vspace{0.2cm}
\noindent \textbf{Theorem D.1 (Convergence of Proxy $Q$ Function)}
Given a fixed policy $\pi$, the proxy $Q$ function $\hat{Q}^{\pi}$ converges to a unique fixed point $\hat{Q}^{\pi}_*(s',a')$ as the number of updates tends to infinity, where $\hat{Q}^{\pi}$ satisfies:
\begin{equation}
\hat{Q}^{\pi}(s,a) = \gamma \mathbb{E}{s' \sim P(\cdot|s,a)} \left[ \max_{a'} \hat{Q}^{\pi}_*(s',a') \right].
\end{equation}
\vspace{0.2cm}
\noindent \textit{Proof.} The convergence of the proxy $Q$ function follows directly from the contraction property of the Bellman operator $\mathcal{T}^{\pi}$ (Lemma D.1) and the Banach fixed-point theorem \citep{puterman2014markov}. Since $\mathcal{T}^{\pi}$ is a contraction mapping, it has a unique fixed point $\hat{Q}^{\pi}_*$, and the sequence ${\hat{Q}^{\pi}_k}{k=0}^{\infty}$ defined by $\hat{Q}^{\pi}_{k+1} = \mathcal{T}^{\pi} \hat{Q}^{\pi}_k$ converges to $\hat{Q}^{\pi}_*$ as $k \to \infty$, regardless of the initial $\hat{Q}^{\pi}_0$

\section{Workflow of Physics-enhanced Reinforcement Learning with Human Framework}
\label{Appendix:Workflow-PE-RLHF}
\begin{algorithm}[H]
\caption{Physics-enhanced Reinforcement Learning with Human Framework (PE-RLHF)}
\begin{algorithmic}[1]
\Require Policy network parameters $\theta$, proxy value function parameters $\phi$, intervention value function parameters $\omega$, and replay buffer $\mathcal{B}$
\While{training is not finished}
\While{episode is not terminated}
\State Observe state $s_t$
\State Sample action $a_t^{\text{AV}} \sim \pi_{\text{AV}}(\cdot \mid s_t; \theta)$
\State $I(s_t, a_t^{\text{AV}}) \gets$ Human mentor determines whether to intervene by observing current state $s_t$
\If{$I(s_t, a_t^{\text{AV}})$ is True}
\State $a_t^{\text{hybrid}} \sim \pi_{\text{hybrid}}(\cdot \mid s_t)$ // Retrieve hybrid intervention action
\State Execute $a_t^{\text{hybrid}}$ within the environment
\If{$I(s_t, a_t^{\text{AV}})$ is True and $I(s_{t-1}, a_{t-1}^{\text{AV}})$ is False}
\State $C^{\text{int}}(s_t, a_t^{\text{AV}}) \gets$ Compute intervention cost following Eq. \ref{eq26}
\Else
\State $C^{\text{int}}(s_t, a_t^{\text{AV}}) \gets 0$ // Set intervention cost to zero
\EndIf
\Else
\State Execute $a_t^{\text{AV}}$ within the environment
\EndIf
\State Observe next state $s_{t+1}$
\State Store transition $(s_t, a_t^{\text{AV}}, a_t^{\text{hybrid}}, I(s_t, a_t^{\text{AV}}), s_{t+1})$ in $\mathcal{B}$
\EndWhile
\For{each gradient step}
\State Sample a mini-batch of transitions $\left(s_t, a_t^{\text{AV}}, a_t^{\text{hybrid}}, I(s_t, a_t^{\text{AV}}), s_{t+1}\right)$ from $\mathcal{B}$
\State Compute target $y$ using Eq. \ref{eq24}
\State Update proxy value function parameters $\phi$ using Eq. \ref{eq25}
\State Update intervention value function parameters $\omega$ using Eq. \ref{eq27}
\State Compute policy gradient $\nabla_{\theta} J(\theta)$ using Eq. \ref{eq28}
\State Update policy parameters $\theta$ using $\nabla_{\theta} J(\theta)$
\EndFor
\EndWhile
\end{algorithmic}
\end{algorithm}

\section{Hyper-parameters}
\label{Appendix:Hyper-parameters}

\begin{table}[H]
\begin{minipage}{0.45\linewidth}
\centering
\caption{PPO/PPO-Lag}
\begin{tabular}{@{}ll@{}}
\toprule
Hyper-parameter             & Value  \\ \midrule
KL Coefficient              & 0.2    \\
$\lambda$ for GAE~\citep{schulman2015high} & 0.95 \\
Discounted Factor $\gamma$   & 0.99  \\
Number of SGD epochs   & 20     \\
Train Batch Size & 4000 \\
SGD mini-batch size & 100 \\
Learning Rate               & 0.00005 \\ 
Clip Parameter $\epsilon$ & 0.2 \\
Cost Limit for PPO-Lag & 1 \\
\bottomrule
\end{tabular}
\end{minipage} \hfill
\begin{minipage}{0.45\linewidth}
\centering
\caption{SAC/SAC-Lag/CQL}
\begin{tabular}{@{}ll@{}}
\toprule
Hyper-parameter             & Value  \\ \midrule
Discounted Factor $\gamma$   & 0.99  \\
$\tau$ for target network update & 0.005 \\
Learning Rate               & 0.0001 \\ 
Environmental horizon $T$ & 1500 \\
Steps before Learning start & 10000\\
Cost Limit for SAC-Lag & 1 \\
BC iterations for CQL & 200000 \\
CQL Loss Temperature $\beta$ & 5 \\
Min Q Weight Multiplier & 0.2 \\
\bottomrule
\end{tabular}
\end{minipage}\hfill
\end{table}

\begin{table}[H]
\begin{minipage}{0.45\linewidth}
\centering
\caption{CPO}
\begin{tabular}{@{}ll@{}}
\toprule
Hyper-parameter             & Value  \\ \midrule
KL Coefficient              & 0.2    \\
$\lambda$ for GAE~\citep{schulman2015high} & 0.95 \\
Discounted Factor $\gamma$   & 0.99  \\
Number of SGD epochs   & 20     \\
Train Batch Size & 8000 \\
SGD mini-batch size & 100 \\
Learning Rate               & 0.00005 \\ 
Clip Parameter $\epsilon$ & 0.2 \\
Cost Limit & 1 \\
\bottomrule
\end{tabular}
\end{minipage}\hfill
\begin{minipage}{0.45\linewidth}
\centering
\caption{BC}
\begin{tabular}{@{}ll@{}}
\toprule
Hyper-parameter             & Value  \\ \midrule
Dataset Size & 49,000 \\
SGD Batch Size & 32 \\
SGD Epoch & 200000 \\
Learning Rate & 0.0001\\
\bottomrule
\end{tabular}
\end{minipage}
\end{table}

\begin{table}[H]
\begin{minipage}{0.45\linewidth}
\centering
\caption{GAIL}
\begin{tabular}{@{}ll@{}}
\toprule
Hyper-parameter             & Value  \\ \midrule
Dataset Size & 49,000 \\
SGD Batch Size & 64 \\
Sample Batch Size &  12800 \\
Generator Learning Rate & 0.0001 \\
Discriminator Learning Rate & 0.005 \\
Generator Optimization Epoch & 5 \\
Discriminator Optimization Epoch & 2000 \\
Clip Parameter $\epsilon$ & 0.2 \\
\bottomrule
\end{tabular}
\end{minipage}\hfill
\begin{minipage}{0.45\linewidth}
\centering
\caption{HG-DAgger}
\begin{tabular}{@{}ll@{}}
\toprule
Hyper-parameter             & Value  \\ 
\midrule
Initializing dataset size & 30,000 \\
Number of data aggregation epoch & 4 \\
Interactions per round & 5000 \\
SGD batch size & 256\\
Learning rate & 0.0004\\
\bottomrule
\end{tabular}
\end{minipage}
\end{table}

\begin{table}[H]
\begin{minipage}{0.45\linewidth}
\centering
\caption{IWR}
\begin{tabular}{@{}ll@{}}
\toprule
Hyper-parameter             & Value  \\ 
\midrule
Initializing dataset size & 30,000 \\
Number of data aggregation epoch & 4 \\
Interactions per round & 5000 \\
SGD batch size & 256\\
Learning rate & 0.0004\\
Re-weight data distribution & True \\
\bottomrule
\end{tabular}
\end{minipage}\hfill
\begin{minipage}{0.45\linewidth}
\centering
\caption{HACO}
\begin{tabular}{@{}ll@{}}
\toprule
Hyper-parameter             & Value  \\ \midrule
Discounted Factor $\gamma$   & 0.99  \\
$\tau$ for Target Network Update & 0.005 \\
Learning Rate               & 0.0001 \\ 
Environmental Horizon $T$ & 1000 \\
Steps before Learning Start & 100\\
Steps per Iteration & 100 \\
Train Batch Size & 1024  \\
CQL Loss Temperature & 10.0 \\
Target Entropy & 2.0\\ 
\bottomrule
\end{tabular}
\end{minipage}
\end{table}

\begin{table}[H]
\centering
\caption{PE-RLHF}
\begin{tabular}{@{}ll|ll@{}}
\toprule
Hyper-parameter    & Value & Hyper-parameter    & Value \\ \midrule
Politeness factor $p^{\text{pol}}$    & 0.1            & Weighting Factor $\psi$               & 1             \\
Acceleration threshold $\Delta a{\text{threshold}}$ & 0.2 & Weighting Factor $\beta$ & 1            \\
Maximum safe deceleration              & 2.0            & Discounted Factor $\gamma$           & 0.99          \\
Lane change time gap                   & 1.0            & $\tau$ for Target Network Update     & 0.005         \\
Maximum acceleration $a_{max}$         & 2.0            & Learning Rate                        & 0.0001        \\
Comfortable braking deceleration $\beta$ & -5.0         & Environmental Horizon               & 1000          \\
Standstill distance $s_0$             & 10.0           & Steps before Learning Start          & 100           \\
Safe time headway $T$                 & 1.5            & Steps per Iteration                  & 100           \\
Exponent for velocity $\eta$          & 4.0            & Train Batch Size                     & 1024          \\
Warmup steps              &      10K        & CQL Loss Temperature       & 10.0          \\ Value takeover threshold $\varepsilon_{\text{select}}$ &    0.5 & Target Entropy      & 2.0    \\ 
\bottomrule
\end{tabular}
\end{table}

\bibliography{mybibfile}

\end{document}